\pdfoutput=1
\documentclass[11pt]{article}

\usepackage{ACL2024}
\usepackage{times}
\usepackage{latexsym}
\usepackage{graphicx}

\usepackage[T1]{fontenc}
\usepackage[utf8]{inputenc}

\usepackage{microtype}
\usepackage{booktabs}
\usepackage{multirow}
\usepackage{makecell}

\usepackage{paralist}
\usepackage{amsmath,amssymb,amsfonts,mathrsfs}

\usepackage[capitalize,noabbrev]{cleveref}

\usepackage[many]{tcolorbox}
\newtcolorbox{myboxgrey}[1]{breakable, colback=gray!5!white,colframe=gray!75!black,fonttitle=\bfseries,title=#1}
\newtcolorbox{myboxred}[1]{breakable, colback=red!5!white,colframe=black,fonttitle=\bfseries,title=#1}
\newtcolorbox{myboxblue}[1]{breakable, colback=blue!5!white,colframe=black,fonttitle=\bfseries,title=#1}
\newtcolorbox{myboxgreen}[1]{breakable, colback=green!5!white,colframe=black,fonttitle=\bfseries,title=#1}
\newtcolorbox{myboxyellow}[1]{breakable,colback=yellow!5!white,colframe=black,fonttitle=\bfseries,title=#1}
\newtcolorbox{myboxgreynotitle}{breakable, colback=gray!5!white,colframe=gray!75!black}

\newcommand{\ours}{\textsc{SparseFit}\xspace}
\newcommand{\textgr}[1]{\textcolor{teal}{#1}}
\newcommand{\textbl}[1]{\textcolor{blue}{#1}}

\usepackage{float}

\usepackage{xspace}
\newcommand*{\eg}{e.g.\@\xspace}
\newcommand*{\ie}{i.e.\@\xspace}

\title{\ours: Few-shot Prompting with Sparse Fine-tuning for Jointly Generating Predictions and Natural Language Explanations}

\author{Jesus Solano \\
  ETH Zürich \\
  {\small \texttt{jesus.solano@inf.ethz.ch}} \\\And
  Mardhiyah Sanni \\
  University of Edinburgh \\
  {\small \texttt{m.o.sanni@sms.ed.ac.uk}} \\ \AND
  Oana-Maria Camburu \\
  University College London \\
  {\small \texttt{o.camburu@ucl.ac.uk}} \\\And
  Pasquale Minervini \\
  University of Edinburgh \\
  {\small \texttt{p.minervini@ed.ac.uk}} \\}

\begin{document}
\maketitle

%%%%%%%%%%%%%%%%%%%%%%%%%%%%%%%%%%%%%%%%%%%%%%%%%%%%%
% ===================================================
% Abstract
% ===================================================
%%%%%%%%%%%%%%%%%%%%%%%%%%%%%%%%%%%%%%%%%%%%%%%%%%%%%

\begin{abstract}

% Abstract Version - LoRA and Prefix-tuning
Models that generate natural language explanations (NLEs) for their predictions have recently gained increasing interest. However, this approach usually demands large datasets of human-written NLEs for the ground-truth answers at training time, which can be expensive and potentially infeasible for some applications. When only a few NLEs are available (a few-shot setup), fine-tuning pre-trained language models (PLMs) in conjunction with prompt-based learning has recently shown promising results. However, PLMs typically have billions of parameters, making full fine-tuning expensive. We propose \ours, a sparse few-shot fine-tuning strategy that leverages discrete prompts to jointly generate predictions and NLEs. We experiment with \ours on three sizes of the T5 language model and four datasets and compare it against existing state-of-the-art Parameter-Efficient Fine-Tuning~(PEFT) techniques. We find that fine-tuning only $6.8\%$ of the model parameters leads to competitive results for both the task performance and the quality of the generated NLEs compared to full fine-tuning of the model and produces better results on average than other PEFT methods in terms of predictive accuracy and NLE quality.

\end{abstract}

%%%%%%%%%%%%%%%%%%%%%%%%%%%%%%%%%%%%%%%%%%%%%%%%%%%%%
% ===================================================
% Introduction
% ===================================================
%%%%%%%%%%%%%%%%%%%%%%%%%%%%%%%%%%%%%%%%%%%%%%%%%%%%%

\section{Introduction}
Despite the tremendous success of neural networks %in Natural Language Processing (NLP)
~\citep{chowdhery2022palm, brown2020gpt3}, these models usually lack human-intelligible explanations for their predictions, which are paramount for ensuring their trustworthiness. 
Building neural models that explain their predictions in natural language has seen increasing interest in recent years~\cite{teachmetoexplain}.
Natural Language Explanations (NLEs) are generally easy to interpret by humans and more expressive than other types of explanations~\citep{wallace2019allennlp, teachmetoexplain}.
However, a significant downside of these models is that they require large datasets of human-written NLEs at training time, which can be expensive and time-consuming to collect. To this end, few-shot learning of NLEs has recently emerged~\cite{marasovic-etal-2022-shot, yordanov2021few}.
However, current techniques involve fine-tuning the \textit{entire} model with a few training NLE examples.
This is computationally expensive since current NLE models can have billions of parameters~\cite{DBLP:journals/cacm/SchwartzDSE20}. %, and can lead to overfitting in a few-shot learning setting~\cite{peters2019tune}.

In this paper, we investigate whether \emph{sparse fine-tuning} (\ie fine-tuning only a subset of parameters), in conjunction with prompt-based learning (\ie, textual instructions provided to a model~\citep{liu2021pre}), can help in scenarios with limited availability of training instances with labels and NLEs.
While sparse fine-tuning has been used in Natural Language Processing~(NLP)~\citep{houlsby2019adapters,logan2021cutting, ben-zaken2022bitfit}, to our knowledge, our work is the first to analyze sparse fine-tuning in the context of jointly generating predictions and NLEs.
We extend the existing sparse fine-tuning strategy that looks only at bias terms~\cite{ben-zaken2022bitfit} to a comprehensive list of all layers and pairs of layers in a language model.
Thus, we propose \ours, an efficient few-shot prompt-based training regime for models generating both predictions and NLEs for their predictions.
%
%We combine prompt-based learning with sparse fine-tuning, which allows us to update only a minimal subset of parameters in the fine-tuning stage.
%
We experiment with \ours on two pre-trained language models (PLMs) that have previously shown high performance on task performance and NLE generation, namely T5~\cite{raffel2020exploring} and \textsc{UnifiedQA}~\cite{dong2019unified}.
%
%Furthermore, we investigate the impact of language model size on the quality of generated explanations.
%
We test our approach on four popular NLE datasets: e-SNLI~\citep{camburu2018snli}, ECQA~\citep{aggarwal2021explanations}, SBIC~\citep{sap-etal-2020-social}, and  ComVE~\citep{wang-etal-2019-make}, and evaluate both the task performance and the quality of the generated NLEs, the latter with both automatic metrics and human evaluation. 
Overall, \ours shows competitive performance in few-shot learning settings with 48 training instances. 
For example, fine-tuning only the Normalization Layer together with the Self-attention Query Layer, which amounts to $6.84\%$ of the model's parameters, consistently gave the best performance (penalized by the number of fine-tuned parameters) on both T5 and \textsc{UnifiedQA} over all four datasets. 
Remarkably, \ours outperforms the current state-of-the-art parameter-efficient fine-tuning~(PEFT) models in terms of both task performance and quality of generated NLEs in two of the four datasets.
Furthermore, we find that fine-tuning other model components that comprise a small fraction of the parameters also consistently leads to competitive results; for instance, the \emph{self-attention query}~(${\sim}6.8\%$ of the parameters), the \emph{self-attention query + LM head}~(${\sim}11.3\%$), and the entire \emph{self-attention layer}~(${\sim}20\%$). %, and the \emph{dense layers}~(${\sim}27\%$).
Moreover, we also applied \ours to larger language models (\ie \texttt{Llama 2-7B}) and found that \ours has competitive performance compared to the best PEFT strategy for all datasets.
Therefore, we conclude that few-shot sparse fine-tuning of PLMs can achieve results competitive with fine-tuning the entire model.

%%%%%%%%%%%%%%%%%%%%%%%%%%%%%%%%%%%%%%%%%%%%%%%%%%%%%
% ===================================================
% Background
% ===================================================
%%%%%%%%%%%%%%%%%%%%%%%%%%%%%%%%%%%%%%%%%%%%%%%%%%%%%
\section{Related Work}

%Deep learning models have shown to be data hungry to be accurate~\cite{hestness2017deephungry,mahajan2018visionhungry}.
%
%This is often challenging when deploying models to real-world scenarios since labeled data is limited for many domains~\cite{wang2020fewshotlearning}. Furthermore, in some particular scenarios, the labeled data is expensive and impossible to gather.
%
Few-shot learning refers to training models with limited labeled data for a given task~\cite{finn2017model,vinyals2016matchingnetworks}.%lampert2009zeroshotlearning}.
It has been successfully applied to several tasks such as image captioning~\citep{dong2018fast}, 
object classification~\citep{ren2018meta}, 
behavioral bio-metrics~\citep{solano2020fewshotbehavioral}, 
% neural network architecture search~\citep{brock2018smash}, 
graph node classification~\citep{satorras2018few}, 
and language modeling~\citep{vinyals2016matchingnetworks}.
Large Language Models~(LLMs) have shown impressive skills to learn in few-shot scenarios~\citep{brown2020gpt3,chowdhery2022palm} thanks to the pre-training corpora size and the statistical capacity of the models~\citep{izacard2022few}.
%
% For example, PaLM~\citep{chowdhery2022palm} has 540B parameters trained
% using the Pathways system~\citep{barham2022pathways} over 780B tokens.
%
%In few-shot learning, authors typically provide a natural language task description, eventually followed by examples telling the model how the task should be completed~\cite{brown2020gpt3}.
%

%%%%%%%%%%%%%%%%%%%%%%
% Parameter Efficient Fine-tuning
%%%%%%%%%%%%%%%%%%%%%%
\paragraph{Parameter-Efficient Fine-Tuning} %\label{sec:background_efficient_finetuning}

Using fine-tuning, LLMs have shown breakthrough language understanding and generation capabilities in a wide range of domains~\citep{raffel2020exploring, brown2020gpt3,chowdhery2022palm}.
%
%Even though the transfer learning of large pre-trained language models has shown impressive performance on many NLP tasks~\citep{radford2018gpt, raffel2020exploring, brown2020gpt3, chowdhery2022palm}, under the traditional fine-tuning paradigm, the entire pre-trained language model is fine-tuned on task-specific supervised data.
%
However, in NLP, the up-stream model (\ie, the model to be fine-tuned) is commonly a LLM with millions of parameters, such as T5~\citep{raffel2020exploring}, BERT~\citep{devlin2019bert}, or GPT-3~\citep{radford2018gpt}, which makes them computationally expensive to fine-tune.
This has led to approaches known in the literature as Parameter-efficient Fine-tuning (PEFT) methods, which fine-tune only a small set of the PLM's parameters or an extra small set of parameters to keep competitive performance in the downstream task.
In this regard, 
% \citet{houlsby2019adapters} developed a strategy that injects model-independent modules, referred to as \emph{adapters}, with only a few trainable parameters.
%
% In their approach, the parameters of the original LM remain fixed while only the parameters of the adapter layers are fine-tuned. 
%
% With a different approach in mind,
\citet{li2021prefixtuning} introduced \emph{Prefix-Tuning}, a strategy that focuses on adding a small task-specific vector to the input so the frozen PLM can adapt its knowledge to further downstream tasks.
\citet{hu2022lora} developed \emph{LoRA}, a technique that 
injects trainable low-rank matrices in the transformer architecture while freezing the pre-trained model weights. \citet{adalora} extended this by proposing \emph{AdaLoRA}, a method that adaptively allocates the rank budget among the low-rank matrices during training according to an importance score.
Later, \citet{ben-zaken2022bitfit} presented BitFit, a novel approach aimed at only fine-tuning the bias terms in each layer of a transformer-based LM.
We extend their work to fine-tuning some layers, or pairs of them, in the LM. 
More recently, \citet{ia3} introduced \emph{(IA)$^3$}, a fine-tuning method that scales the intermediate activations in a model with learned vectors.
%
% Later, \citet{guo2021parameter} developed \emph{diff-pruning}, a strategy that focuses on adding sparse task-specific difference-vector to the original parameters in the PLM.
%

%%%%%%%%%%%%%%%%%%%%%%
% Explainability of Neural Models
%%%%%%%%%%%%%%%%%%%%%%

\paragraph{Explainability of Neural Models} %\label{sec:background_explainability}

%Neural networks have shown impressive accuracy in many tasks such as machine translation, abstractive summarization, text classification, and question answering, among other tasks~\citep{chowdhery2022palm,brown2020gpt3,khashabi2020unifiedqa}.
%
%However, given that those models are composed of several neural blocks, and each block is composed of multiple layers, an explanation of the model's behavior is hard to build.
%
%There is no general interpretation of the relationship between the input features and their prediction~\citep{marasovic-etal-2022-shot}.
%
%Consequently, neural models are often referred to as \emph{black-boxes}~\citep{alvarez2017causalblackbox}.
%
Several approaches have been proposed in the literature to bring a degree of explainability to the predictions of neural models, using %~\citep{camburu2020explaining}.
%
%There are many 
different forms of explanations, such as
\begin{inparaenum}[(1)]
\item Feature-based explanations~\citep{ribeiro2016lime, 
% lundberg2017unified,
shrikumar2017learning, yoon2018invase, leisha},
\item Natural Language Explanations~\citep{camburu2018snli,nmarasovic2020natural, mimicnle, rexc},
\item Counterfactual explanations~\citep{akula2020cocox}, and
\item Surrogate explanations~\citep{alaa2019surrogate}.
\end{inparaenum}
In this paper, we focus on models that provide NLEs, %a natural language explanation for different NLP tasks, such as natural language inference, offensive content detection, and common-sense question answering.
%\paragraph{Natural Language Explanations}%\label{sec:background_natural_language_explanations}
%
%NLEs are textual expressions that justify the predictions of a given neural model.
%
%These explanations are commonly 
i.e., free-form text stating the reasons behind a prediction.
Being in natural language, NLEs should be easy to interpret by humans and more expressive than other types of explanations, as they can %capture a conceptual representation of the target model inference process
present arguments that are not present in the input~\citep{teachmetoexplain, struggles, kaur}.
NLEs have been applied to several domains such as question answering~\citep{narang2020wt5}, natural language inference~\citep{camburu2018snli}, recommendation systems~\citep{chen2021generate}, reinforcement learning~\citep{ehsan2018rationalization}, medical imagining \citep{mimicnle}, visual-textual reasoning \citep{%park2018multimodal,
Hendricks_2018_ECCV, kayser2021vil, rexc}, 
% self-driving cars \citep{kim2018textual},
and solving mathematical problems~\citep{ling2017program}.

To make neural models capable of generating accurate NLEs, the most common approach consists of annotating predictions with human-written explanations and training models to generate the NLEs by casting them as a sequence generation task~\citep{camburu2018snli}.
However, gathering large datasets with human-written NLEs is expensive and time-consuming. 
To address this, \citet{yordanov2021few} proposed a vanilla transfer learning strategy to learn from a few NLEs but abundant labels in a task by fine-tuning a PLM trained on a vast number of NLEs from other domains.
%
% Their results show that generating out-of-domain NLEs via transfer learning outperforms single-task training from scratch for NLE generation.
%
More recently, \citet{marasovic-etal-2022-shot} introduced the FEB benchmark for few-shot learning of NLEs and a prompt-based fine-tuning strategy, which we use as a baseline in our work.
%
% \citet{yordanov2021few} proposed a vanilla transfer learning strategy to learn from a few NLEs but abundant labels in a task (child task), given that the model was pre-trained over a vast number of NLEs belonging to a different task in another domain (parent task).
%

%

%%%%%%%%%%%%%%%%%%%%%%%%%%%%%%%%%%%%%%%%%%%%%%%%%%%%%
% ===================================================
% Approach
% ===================================================
%%%%%%%%%%%%%%%%%%%%%%%%%%%%%%%%%%%%%%%%%%%%%%%%%%%%%
\vspace{0.3cm}

\section{\ours}

% Few-shot learning can also be accomplished by combining textual templates (referred to as \emph{prompts}) with fine-tuning~\cite{izacard2022few}.
% %
% This strategy has proved effective in various NLP tasks, such as classification~\citep{gao2021making, schick2021exploiting} or generation~\citep{radford2019language, schick2021shot}.
% %
% The main idea behind textual prompts is to reformulate the downstream task to look more familiar to the pre-training objective.
% %
% One of those reformulation techniques is adding text snippets in the original text input (\emph{prompt-engineering}), so the language model can recognize the context of the downstream task~\cite{schick2021shot}.
% %
% Recently, \citet{marasovic-etal-2022-shot} explored the idea of training models that jointly generate predictions and explanations -- referred to as \emph{self-rationalization} models~\citep{marasovic-etal-2022-shot,wiegreffe-etal-2021-measuring} -- in few-shot scenarios.
% %
% The main goal of their study was to investigate whether natural language prompts could be used, in addition to model fine-tuning, to induce a self-rationalization behavior in large pre-trained language models.
% %
% Even though they built their idea on top of moderately small-sized language models based on T5~\citep{raffel2020exploring}, their approach fine-tunes the entire set of parameters, which could be computationally expensive and potentially lead to sub-optimal results, especially in few-shots learning settings~\cite{peters2019tune}.
%
We propose \ours, an efficient few-shot NLE training strategy that focuses on fine-tuning only a subset of parameters in a large LM. \ours is inspired by (1)~\citet{marasovic-etal-2022-shot}, who used fine-tuning and prompts to do few-shot learning of labels and NLEs; and (2)~BitFit \citet{ben-zaken2022bitfit}, who showed that fine-tuning only the bias terms in a PLM leads to competitive (and sometimes better) performance than fine-tuning the entire model.
We extend BitFit by exploring the fine-tuning of different components (\ie, layers or blocks) %and certain pairs of them 
in the PLM's architecture. In particular, we study the self-rationalization performance after fine-tuning the following components in the T5 model:
\begin{inparaenum}[(1)]
\item the encoder blocks,
\item the decoder blocks, 
\item the language model head,
\item the self-attention layers,
\item the feed-forward networks,  
\item the normalization layer, and
\item all pairs of the above components that do not contain the encoder and decoder~(see Appendix~\ref{ap:sparse_full_results}).
\end{inparaenum}
Given that \textsc{UnifiedQA} model's architecture is the same as the one in T5, the interpretation of active parameters holds for \textsc{UnifiedQA}. 

%Our aim is to find a general guidance on which component, or pair of components, should be fine-tuned to achieve competitive performance while updating a minimum number of parameters. 
%
We aim to identify a set of guidelines for identifying which components should be fine-tuned to achieve competitive performance while updating a minimum number of parameters. 
Notice that when fine-tuning any component, or pair of components, we freeze all other PLM's parameters and train the LM to conditionally generate a text in the form of \textit{``[label] because [explanation]''}.
%
% Notice that we freeze all other LM's parameters that do not belongs to the component to be fine-tuned, and train the LM to conditionally generate text in the form of \textit{``[label] because [explanation]''}.

% Encoder Tuning
\paragraph{Encoder} \label{sec:sparse_encoder}
The T5 encoder comprises $N$ transformer blocks, each composed of three layers: self-attention, position-wise fully connected layer, and layer normalization.
%
% The number of blocks depends on the size of the T5 model ($N=12$ blocks for \texttt{T5-base}, $N=24$ for \texttt{T5-large}, and $N=36$ for \texttt{T5-3B}). The encoder accounts for roughly $41\%$ of the model parameters.
The number of blocks depends on the T5 variant~(12 blocks for \texttt{T5-base}, 24 for \texttt{T5-large}, and 36 for \texttt{T5-3B}). 
The encoder accounts for roughly $41\%$ of the model parameters.
%
%When fine-tuning the encoder, we freeze all the parameters outside its $N$ blocks.
%
%After freezing the rest of the model, the parameters that will be updated when fine-tuning the encoder is roughly $41\%$ of the model parameters.
%

% Decoder Tuning
\paragraph{Decoder}\label{sec:sparse_decoder}
The decoder accounts for roughly $54\%$ of T5 model parameters.
In addition to the self-attention, fully connected layer, and layer normalization, it also includes an encoder-decoder attention layer in its blocks, which we fine-tune as part of fine-tuning the decoder.
%
%When fine-tuning the decoder using \ours, all the parameters outside the decoder blocks are frozen, including the language model head.
%
%Note that all the parameters in the decoder blocks are fine-tuned, including the encoder-decoder attention layers.

% Blocks Tuning
\paragraph{LM Head}\label{sec:sparse_lmhead}

On top of the decoder, T5 has a language modeling head for generating text based on the corpus.
The LM head accounts for roughly $5\%$ of total model parameters.
%$ 

% Attention Tuning
\paragraph{Attention Layer}\label{sec:sparse_attention}
Each of the transformer blocks starts with a self-attention layer.
There are three types of parameters in the self-attention layer, namely, for computing the \emph{query matrix} $Q$, the \emph{key matrix} $K$, and the \emph{value matrix} $V$. We propose to explore the fine-tuning of each self-attention matrix as a possible \ours configuration. We also explore fine-tuning the \emph{entire Self-attention Layer}~($Q$,~$K$,~$V$).
On average, the percentage of trainable parameters associated with each matrix accounts for roughly $6\%$ of model parameters. Note that the attention parameters in the encoder-decoder attention are not updated in %any 
this setting (they are only updated together with the decoder).

% Normalization Layer Tuning
\paragraph{Layer Normalization}\label{sec:sparse_norm}
The normalization layers are intended to improve the training speed of the models~\cite{ba2016layernorm}.
The T5 model includes two \textit{Layer Normalization} components per block, one after the self-attention layer and one after the feed-forwards networks.
Unlike the original paper for layer normalization~\citep{ba2016layernorm}, the T5 model uses a simplified version of the layer normalization that only re-scales the activations. % (no bias is applied).
The percentage of learnable weights in the layer normalization is roughly $0.2\%$ of the parameters. % for different T5/\textsc{UnifiedQA} model sizes.

\vspace{0.3cm}

%%%%%%%%%%%%%%%%%%%%%%%%%%%%%%%%%%%%%%%%%%%%%%%%%%%%%
% ===================================================
% Experimental Setup
% ===================================================
%%%%%%%%%%%%%%%%%%%%%%%%%%%%%%%%%%%%%%%%%%%%%%%%%%%%%
\section{Experiments}

% Datasets
\paragraph{Datasets}\label{sec:datasets}
We follow the FEB benchmark for few-shot learning of NLEs~\citep{marasovic-etal-2022-shot} and consider four NLE datasets: e-SNLI for natural language inference~\citep{camburu2018snli}, ECQA for multiple-choice question answering~\citep{aggarwal2021explanations}, ComVE for commonsense classification~\citep{wang-etal-2019-make}, and SBIC for offensiveness classification~\citep{sap-etal-2020-social}.
%

% asdfasdf

% Downstream, NLE quality and Compound evaluation summary (T5-large)
\begin{table*}[t]
    \centering
    \resizebox{0.9\textwidth}{!}{
    \begin{tabular}{c c c c c c c}
    \toprule
        %\textbf{Dataset} & \multicolumn{4}{c| }{Dataset} & \textbf{Avg} \\ \hline \hline
        % \midrule
        \textbf{\ours} & & ComVE & ECQA & SBIC & e-SNLI &  \textbf{Avg} \\ %\hline
        \midrule
        \multirow{2}{*}{\shortstack{Baseline \\ ($100.00\%$)}} & Acc. & \textbf{80.5} {\tiny $\pm4.5$} &  57.6 {\tiny $\pm2.6$} &   \textbf{70.1} {\tiny $\pm3.4$} &   84.8 {\tiny $\pm2.6$} &  73.3 {\tiny $\pm3.3$} \\
                                            & nBERTs &   \textbf{74.2} {\tiny $\pm4.2$} &  51.7 {\tiny $\pm2.4$} &   \textbf{67.8} {\tiny $\pm3.3$} &  76.9 {\tiny $\pm2.5$} &  67.7 {\tiny $\pm3.1$} \\
                                            % & Comp. &  \textbf{60.0} {\tiny $\pm6.4$} &  29.8 {\tiny $\pm2.7$} &  \textbf{47.7} {\tiny $\pm4.6$} &  65.3 {\tiny $\pm4.0$} &  50.7 {\tiny $\pm4.5$} \\
                                            \hline
        \multirow{2}{*}{\shortstack{Decoder \\ ($54.60\%$)}} & Acc. &  67.3 {\tiny $\pm6.0$}  $\triangledown$ &     58.5 {\tiny $\pm2.6$} &   66.8 {\tiny $\pm3.1$} $\triangledown$ &   86.6 {\tiny $\pm1.7$} $\triangledown$ &  69.8 {\tiny $\pm3.4$} \\
                                            & nBERTs &   61.7 {\tiny $\pm5.5$} $\triangledown$ &  \textbf{52.3} {\tiny $\pm2.4$} $\triangledown$ &   64.7 {\tiny $\pm2.7$} &  \textbf{78.3} {\tiny $\pm1.6$} $\triangledown$ &  64.3 {\tiny $\pm3.0$} \\
                                            % & Comp. &  41.9 {\tiny $\pm7.5$} &  \textbf{30.6} {\tiny $\pm2.8$} &  43.3 {\tiny $\pm3.7$} &  \textbf{67.8} {\tiny $\pm2.7$} &  45.9 {\tiny $\pm4.2$} \\
                                            \hline
        \multirow{2}{*}{\shortstack{Encoder \\ ($40.95\%$)}} & Acc. & 72.6 {\tiny $\pm6.1$} $\triangledown$ &  53.2 {\tiny $\pm3.6$} $\triangledown$ &   62.4 {\tiny $\pm6.5$} $\triangledown$ &   79.0 {\tiny $\pm3.4$} $\triangledown$ &  66.8 {\tiny $\pm4.9$} \\
                                            & nBERTs &   67.1 {\tiny $\pm5.7$} &  47.2 {\tiny $\pm3.2$} $\triangledown$ &   58.7 {\tiny $\pm6.5$} $\triangledown$ &  72.4 {\tiny $\pm3.2$} $\triangledown$ &  61.3 {\tiny $\pm4.6$} \\
                                            % & Comp. &  49.0 {\tiny $\pm8.1$} &  25.2 {\tiny $\pm3.4$} &  37.1 {\tiny $\pm7.8$} &  57.2 {\tiny $\pm4.9$} &  42.1 {\tiny $\pm6.0$} \\
                                            \hline
        \multirow{2}{*}{\shortstack{Dense.wo \\ ($27.29\%$)}} & Acc. &  61.3 {\tiny $\pm4.4$} $\triangledown$ &  56.1 {\tiny $\pm2.1$} $\triangledown$ &   62.4 {\tiny $\pm2.6$} $\triangledown$ &   84.0 {\tiny $\pm1.9$} &  65.9 {\tiny $\pm2.8$} \\
                                            & nBERTs &   56.4 {\tiny $\pm4.1$} $\triangledown$ &   0.0 {\tiny $\pm0.0$} $\triangledown$ &   59.8 {\tiny $\pm2.6$} $\triangledown$ &  74.7 {\tiny $\pm2.6$} $\triangledown$ &  47.7 {\tiny $\pm2.3$} \\
                                            % & Comp. &  34.8 {\tiny $\pm5.0$} &   0.0 {\tiny $\pm0.0$} &  37.3 {\tiny $\pm3.3$} &  62.8 {\tiny $\pm3.4$} &  33.7 {\tiny $\pm2.9$} \\
                                            \hline
        % \multirow{2}{*}{\shortstack{Dense.wi \\ ($27.29\%$)}} & Acc. &  54.2 {\tiny $\pm2.5$} $\triangledown$ &  57.0 {\tiny $\pm2.4$} &   23.2 {\tiny $\pm8.9$} $\triangledown$ &   56.2 {\tiny $\pm5.5$} $\triangledown$ &  47.7 {\tiny $\pm4.8$} \\
        %                                     & nBERTs &   49.8 {\tiny $\pm2.3$} $\triangledown$ &   0.0 {\tiny $\pm0.0$} $\triangledown$ &   23.2 {\tiny $\pm8.9$} $\triangledown$ &  14.3 {\tiny $\pm3.6$} $\triangledown$ &  21.8 {\tiny $\pm3.7$} \\
        %                                     % & Comp. &  27.0 {\tiny $\pm2.6$} &   0.0 {\tiny $\pm0.0$} &   6.2 {\tiny $\pm3.4$} &   8.2 {\tiny $\pm2.5$} &  10.3 {\tiny $\pm2.1$} \\
        %                                     \hline
        \multirow{2}{*}{\shortstack{Self-attention~(KQV) \\ ($20.47\%$)}} & Acc. &  \textbl{76.2} {\tiny $\pm4.4$} $\triangledown$ &  56.9 {\tiny $\pm3.0$} &   \textbl{69.9} {\tiny $\pm3.8$} &   83.3 {\tiny $\pm2.4$} $\triangledown$ &  \textbl{71.6} {\tiny $\pm3.4$}  \\
                                            & nBERTs &   \textbl{70.3} {\tiny $\pm4.0$} $\triangledown$ &  50.2 {\tiny $\pm2.7$} $\triangledown$ &   \textbl{67.4} {\tiny $\pm3.9$} $\triangledown$ &  76.1 {\tiny $\pm2.2$} $\triangledown$ &  \textbl{66.0} {\tiny $\pm3.2$} \\
                                            % & Comp. &  \textbl{53.8} {\tiny $\pm6.0$} &  28.6 {\tiny $\pm3.0$} &  \textbl{47.2} {\tiny $\pm5.2$} &  63.4 {\tiny $\pm3.7$} &  \textbl{48.3} {\tiny $\pm4.5$} \\
                                            \hline
        \multirow{2}{*}{\shortstack{LM head +  Attention.Q \\ ($11.28\%$)}} & Acc. &  74.8 {\tiny $\pm5.0$} $\triangledown$ &  55.4 {\tiny $\pm2.7$} $\triangledown$ &   67.1 {\tiny $\pm5.2$} $\triangledown$ &   \textgr{82.8} {\tiny $\pm3.0$} $\triangledown$ &  70.0 {\tiny $\pm4.0$} \\
                                            & nBERTs &   69.0 {\tiny $\pm4.6$} &  43.7 {\tiny $\pm4.3$} $\triangledown$ &   64.5 {\tiny $\pm5.5$}  &  \textgr{75.8} {\tiny $\pm2.8$} $\triangledown$ &  63.2 {\tiny $\pm4.3$} \\
                                            % & Comp. &  51.8 {\tiny $\pm6.6$} &  24.3 {\tiny $\pm3.3$} &  43.5 {\tiny $\pm6.9$} &  62.9 {\tiny $\pm4.5$} &  45.6 {\tiny $\pm5.3$} \\
                                            \hline
        \multirow{2}{*}{\shortstack{LM head \\ ($4.46\%$)}} & Acc. & 15.6 {\tiny $\pm1.3$} $\triangledown$ &  58.9 {\tiny $\pm2.3$} $\triangledown$ &    0.2 {\tiny $\pm0.2$} $\triangledown$ &   \textbf{86.7} {\tiny $\pm1.8$} $\triangledown$ &  40.3 {\tiny $\pm1.4$} \\
                                            & nBERTs &    0.0 {\tiny $\pm0.0$} $\triangledown$ &   0.0 {\tiny $\pm0.0$} $\triangledown$ &    0.2 {\tiny $\pm0.2$} $\triangledown$ &   0.0 {\tiny $\pm0.0$} $\triangledown$ &   0.0 {\tiny $\pm0.0$} \\
                                            % & Comp. &   0.0 {\tiny $\pm0.0$} &   0.0 {\tiny $\pm0.0$} &   0.0 {\tiny $\pm0.0$} &   0.0 {\tiny $\pm0.0$} &   0.0 {\tiny $\pm0.0$} \\
                                            \hline
        \multirow{2}{*}{\shortstack{LayerNorm +  Attention.Q \\ ($6.84\%$)}} & Acc. & \textgr{74.9} {\tiny $\pm5.3$} $\triangledown$ &  \textgr{55.8} {\tiny $\pm3.1$} $\triangledown$ &   \textgr{67.0} {\tiny $\pm4.4$} $\triangledown$&   82.6 {\tiny $\pm2.7$} $\triangledown$ &  \textgr{70.1} {\tiny $\pm3.9$} \\
                                            & nBERTs &   \textgr{69.0} {\tiny $\pm4.8$} &  \textgr{45.9} {\tiny $\pm3.7$} $\triangledown$ &   \textgr{64.3} {\tiny $\pm4.7$} &  75.6 {\tiny $\pm2.5$} $\triangledown$ &  \textgr{63.7} {\tiny $\pm3.9$} \\
                                            % & Comp. &  \textgr{51.9} {\tiny $\pm7.1$} &  \textgr{25.7} {\tiny $\pm3.2$} &  \textgr{43.3} {\tiny $\pm5.9$} &  62.6 {\tiny $\pm4.1$} &  \textgr{45.9} {\tiny $\pm5.1$} \\
                                            \hline
        \multirow{2}{*}{\shortstack{Attention.K \\ ($6.82\%$)}} & Acc. & 48.8 {\tiny $\pm2.8$} $\triangledown$ &  56.7 {\tiny $\pm2.5$} $\triangledown$ &    0.2 {\tiny $\pm0.2$} $\triangledown$ &  19.6 {\tiny $\pm11.5$} $\triangledown$ &  31.3 {\tiny $\pm4.3$} \\
                                            & nBERTs &  19.4 {\tiny $\pm10.0$} $\triangledown$ &   0.0 {\tiny $\pm0.0$} $\triangledown$ &    0.1 {\tiny $\pm0.2$} $\triangledown$ &   0.2 {\tiny $\pm0.3$} $\triangledown$ &   4.9 {\tiny $\pm2.6$} \\
                                            % & Comp. &   9.6 {\tiny $\pm5.1$} &   0.0 {\tiny $\pm0.0$} &   0.0 {\tiny $\pm0.0$} &   0.0 {\tiny $\pm0.1$} &   2.4 {\tiny $\pm1.3$} \\
                                            \hline
        \multirow{2}{*}{\shortstack{Attention.Q \\ ($6.82\%$)}} & Acc. & 74.8 {\tiny $\pm5.1$} $\triangledown$ &  55.5 {\tiny $\pm3.2$} $\triangledown$ &   66.9 {\tiny $\pm4.6$} $\triangledown$ &   \textgr{82.8} {\tiny $\pm2.6$} $\triangledown$ &  70.0 {\tiny $\pm3.8$} \\
                                            & nBERTs &  68.9 {\tiny $\pm4.7$} &  43.4 {\tiny $\pm4.8$} $\triangledown$ &   64.2 {\tiny $\pm4.8$} &  \textgr{75.8} {\tiny $\pm2.3$} $\triangledown$ &  63.1 {\tiny $\pm4.2$} \\
                                            % & Comp. &  51.8 {\tiny $\pm6.8$} &  24.2 {\tiny $\pm3.6$} &  43.2 {\tiny $\pm6.1$} &  \textgr{62.8} {\tiny $\pm3.8$} &  45.5 {\tiny $\pm5.1$} \\
                                            \hline
        \multirow{2}{*}{\shortstack{Attention.V \\ ($6.82\%$)}} & Acc. & 55.5 {\tiny $\pm3.0$} $\triangledown$ &  53.1 {\tiny $\pm2.8$} $\triangledown$ &  30.1 {\tiny $\pm10.2$} $\triangledown$ &   84.2 {\tiny $\pm2.0$} &  55.7 {\tiny $\pm4.5$} \\
                                            & nBERTs &   51.0 {\tiny $\pm2.8$} $\triangledown$ &   0.0 {\tiny $\pm0.0$} $\triangledown$&  30.1 {\tiny $\pm10.2$} $\triangledown$ &  71.7 {\tiny $\pm3.4$} $\triangledown$ &  38.2 {\tiny $\pm4.1$} \\
                                            % & Comp. &  28.4 {\tiny $\pm3.2$} &   0.0 {\tiny $\pm0.0$} &  10.1 {\tiny $\pm5.4$} &  60.4 {\tiny $\pm3.6$} &  24.7 {\tiny $\pm3.0$} \\
                                            \hline
        \multirow{2}{*}{\shortstack{LayerNorm \\ ($0.02\%$)}} & Acc. & 34.3 {\tiny $\pm2.4$} $\triangledown$ &  \textbf{59.0} {\tiny $\pm2.4$} $\triangledown$ &    0.3 {\tiny $\pm0.3$} $\triangledown$ &   86.6{\tiny $\pm1.8$} $\triangledown$ &  45.0 {\tiny $\pm1.7$}  \\
                                            & nBERTs &    0.0 {\tiny $\pm0.0$} $\triangledown$ &   0.0 {\tiny $\pm0.0$} $\triangledown$ &    0.2 {\tiny $\pm0.2$} $\triangledown$ &   0.0 {\tiny $\pm0.0$} $\triangledown$ &   0.1 {\tiny $\pm0.1$} \\
                                            % & Comp. &   0.0 {\tiny $\pm0.0$} &   0.0 {\tiny $\pm0.0$} &   0.0 {\tiny $\pm0.0$} &   0.0 {\tiny $\pm0.0$} &   0.0 {\tiny $\pm0.0$} \\
    \bottomrule
    \end{tabular}
    
    }
    \caption{Summary of best performing \ours configurations for \texttt{T5-large}. We report the average and the standard deviation over the 60 few-shot train-validation splits for the \textbf{accuracy} metric and the normalized BERTScore~(\textbf{nBERTs}).    
    % and \textbf{Compound Metric}~(Acc $\times$ BERTScore) metric. 
    In brackets are the percentages of fine-tuned weights for each \ours configuration. We show in \textbf{bold} the setting with the highest metric for each dataset, in \textbl{blue} the highest performance among \ours without considering the number of parameters, and in \textgr{green} the best-performing setting after considering the percentage of fine-tuned parameters. The trade-off between parameters and performances was computed using~$(1 - \% \text{params}) \times $~nBERTs). Significance testing was assessed via mean t-test compared with the baseline:~$\triangledown$~represents a p-value lower than $10^{-2}$.}
    \label{tab:sparse_summary_all}
\end{table*}

% Few-shot learning splits
\paragraph{Few-shot Learning Data Splits}%\label{sec:few_shot_split}
We also follow the few-shot evaluation protocol used by \citet{marasovic-etal-2022-shot}.
We use their 60 train-validation splits to run our experiments. Each experiment is run with 48 examples in the training set and 350 examples in the validation set.
Note that, depending on the dataset, the number of training examples per label varies: e-SNLI has 16 examples for each label, ECQA 48, SBIC 24, and  omVE 24, resulting in 48 training examples for all datasets.
%
% We implemented their approach use the same 60 splits as \citet{marasovic-etal-2022-shot} to compare our results directly comparable against theirs. 

% Explanations t5-large all Datasets 
\begin{figure*}[t]
    \centering
    \includegraphics[width=0.95\textwidth]{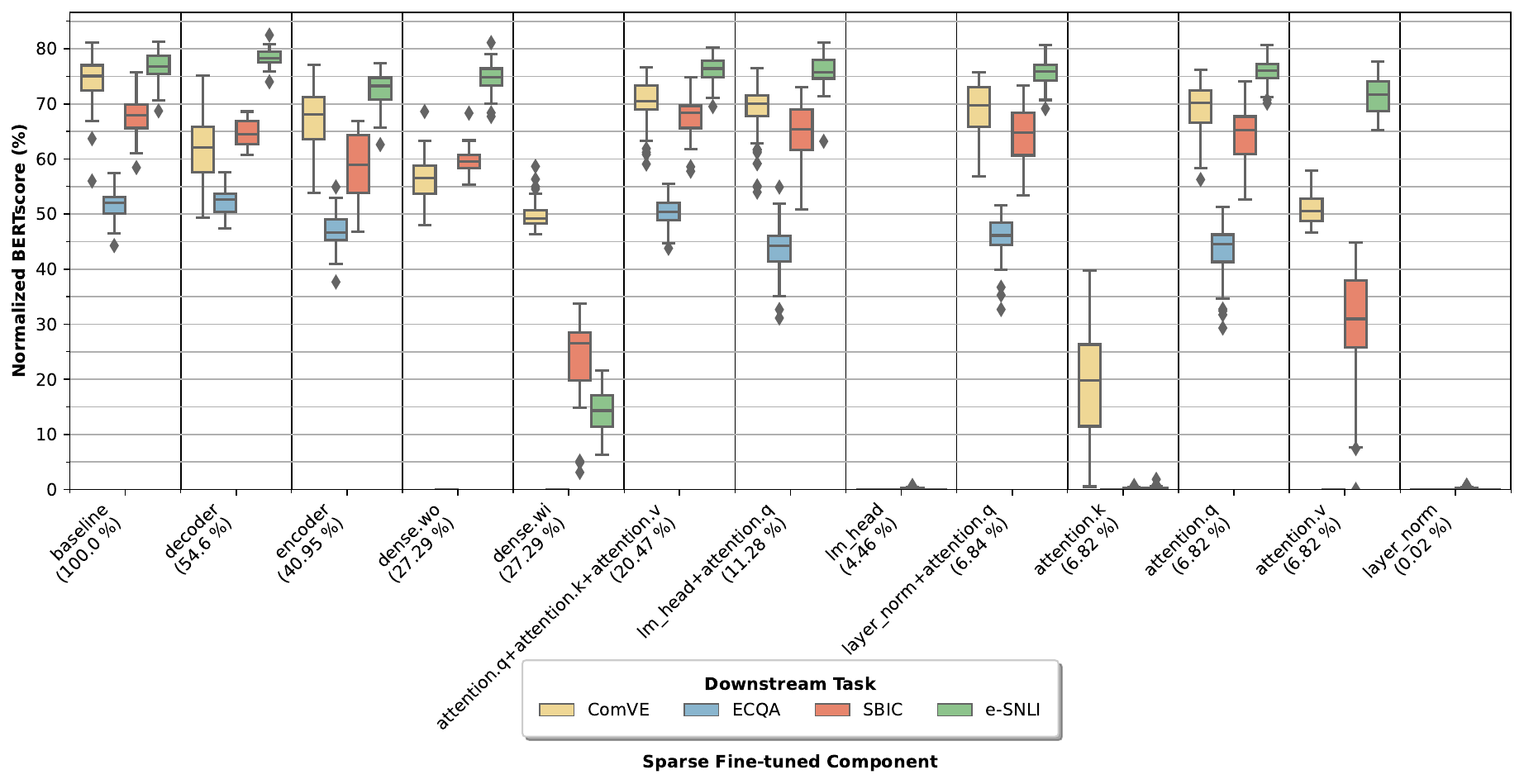}
    \caption{Distribution of the \textbf{normalized BERTScore} for different \ours settings of sparse fine-tuning for \texttt{T5-large}. The percentage of fine-tuned parameters is shown between brackets.}
    \label{fig:all_t5large_explanations}
\end{figure*}

% Training procedure
% \vspace{-1em}
\paragraph{Training Procedure} %\label{sec:training_procedure}
Following \citet{marasovic-etal-2022-shot}, we fine-tune T5~\citep{raffel2020exploring} and \textsc{UnifiedQA}~\citep{khashabi2020unifiedqa}.
Depending on the setup, the gradients are activated for specific parameters (\ours) or the entire model (baseline).
We report our experimental results for the baseline, and observe a consistent behavior with the one reported by \citet{marasovic-etal-2022-shot}.
Additionally, to compare with other PEFT baselines, we adapted LoRA~\cite{hu2022lora}, AdaLoRA~\cite{adalora} and (IA)$^3$~\cite{ia3}.
% and Prefix-Tuning~\cite{li2021prefixtuning} for NLEs.
%
%As an additional PEFT baseline, we adapt the model using a parallel application of LoRA with Prefix-Tuning~\cite{li2021prefixtuning} due to the orthogonality of both these methods.
%
We use the PEFT implementation developed by Hugging Face~\cite{peft}. 
In our implementation of (IA)$^3$, we deviate slightly from its original implementation since we learn scaling vectors for all layers in the model instead of learning them only for the attention modules. This resulted in a significant performance increase.
%
% Also, this implementation approach means we are fine-tuning a maximal number of parameters with (IA)$^3$.
% on two of the four datasets, without adversely affecting the performance on the other two. 
%
For the \ours configurations, we fine-tune each component (or pair) for 25 epochs with a batch size of 4 samples.
Similarly to \citet{marasovic-etal-2022-shot}, we use the AdamW optimizer~\citep{loshchilov2019adamw} with a fixed learning rate of $0.00003$.
%
% The inference is performed using conditional text generation on the validation set.
Conditional text generation is used to do the inference on the validation set.
%
% During the generation process, we only constrain the maximum length of the generated NLE. 
% 
% Training and evaluation take, on average, $23.2$ minutes, running on an   NVIDIA P100.
%
% 
Training and evaluation were run on an NVIDIA P100, and took $23.2${\small min}, on average.
%
% Our code is available at \url{https://github.com/SanniM3/predicitons_with_explanations}
%
\footnote{Our code is available at \url{https://github.com/SanniM3/predicitons_with_explanations}}

% Automatic Evaluation
% \vspace{-1em}
\paragraph{Automatic Evaluation} %\label{sec:automatic_evaluation}
The evaluation considers the task accuracy and the quality of the generated NLEs. 
To automatically evaluate the quality of the NLEs, we follow \citet{marasovic-etal-2022-shot} and use the BERTScore~\citep{zhang2019bertscore}, which was shown by \citet{kayser2021vil} to correlate best with human evaluation in NLEs. 
%we consider metrics for natural language generation (NLG) such as BLEU~\citep{papineni2002bleu}, Rouge-L~\citep{lin2004automatic}, METEOR~\citep{banerjee2005meteor}, and BERTScore~\citep{zhang2019bertscore}. 
%
%Recently, \citet{kayser2021vil} analyzed the correlation between human judgment and automatic NLG metrics for NLEs, finding that BERTScore and METEOR are the metrics that exhibit the highest correlation with human judgments.
%
%Given this and following \citet{marasovic-etal-2022-shot}, we also use the BERTScore. % for evaluating generated natural language explanations.
%
As in \citet{marasovic-etal-2022-shot}, we compute a \textbf{normalized BERTScore} that assigns a zero score to empty NLEs, or NLEs of incorrectly predicted samples (since one would not expect, nor want, an NLE to be plausible if the prediction was wrong).
We report the averages and standard deviations of the accuracy and the normalized BERTScore over the 60 train-validation splits for each fine-tuning configuration. %to measure the generalization properties of different configurations.
%

% Human Evaluation
\paragraph{Human Evaluation} %\label{sec:human_evaluation}
In addition to the normalized BERTScore, we perform a smaller-scale human evaluation to assess the quality of NLEs for the best-performing \ours configurations. 
We use the NLEs associated with the first 30 correctly predicted samples (balanced to the number of classes) in each validation set for human evaluation.
%
% To make the evaluation more robust, the 30 samples are chosen to balance the number of classes. 
%
Our human evaluation framework follows those of \citet{kayser2021vil,marasovic-etal-2022-shot}.
For the NLE quality assessment, each annotator is asked to answer the question: \emph{``Does the explanation justify the answer?"} and select one of four possible answers: \emph{yes}, \emph{weak yes}, \emph{weak no}, or \emph{no}.
Moreover, we also ask the annotators to identify the main shortcomings, if any, of the generated NLEs.
The possible shortcomings categories are 
\begin{inparaenum}[(1)]
    \item nonsensical, 
    \item contradictory, 
    \item lack of explanation, 
    \item incomplete explanation, 
    \item input repetition, 
    \item hallucination, 
    \item extra words at the end, 
    \item true but uncorrelated, 
    \item inaccurate, and 
    \item one word.
\end{inparaenum}
An author and a third-party annotator 
% (with an MSc degree in Economics) 
performed independent annotations of the whole set of NLEs chosen to be evaluated (600 examples in total).
As in \citet{kayser2021vil}, we compute a numerical value for the quality of the NLEs by mapping the four answers as follows: $\text{yes} \xrightarrow{} 1$, $\text{weak yes} \xrightarrow{} \frac{2}{3}$, $\text{weak no} \xrightarrow{} \frac{1}{3}$, and $\text{no} \xrightarrow{} 0$; and averaging over all annotations per model.
%

%%%%%%%%%%%%%%%%%%%%%%%%%%%%%%%%%%%%%%%%%%%%%%%%%%%%%
% ===================================================
% Results
% ===================================================
%%%%%%%%%%%%%%%%%%%%%%%%%%%%%%%%%%%%%%%%%%%%%%%%%%%%%

%
\subsection{Results}
To evaluate \ours, we compute the task accuracy and the quality of the generated NLEs.
Given that there are 62 possible configurations (single layers plus pairs of layers), for space reasons, the following shows the best configurations based on the model's generalization properties. 
The results for all configurations are shown in \cref{ap:sparse_full_results}. 
%

% Table for comparison with Parameter-Efficient Fine-Tuning (PEFT) strategies
\begin{table*}[t]
    \centering
    \resizebox{\textwidth}{!}{
    
    \begin{tabular}{c c c c c c c c}
    \toprule
        %\textbf{Dataset} & \multicolumn{4}{c| }{Dataset} & \textbf{Avg} \\ \hline \hline
        % \midrule
         & \bf FLOPS & & \bf ComVE & \bf ECQA & \bf SBIC & \bf e-SNLI &  \textbf{Avg} \\ %\hline
        \midrule
        %         \multirow{2}{*}{Baseline} & Acc. & 80.5 {\tiny $\pm4.5$} &  57.6 {\tiny $\pm2.6$} &   70.1 {\tiny $\pm3.4$} &   84.8 {\tiny $\pm2.6$} &  73.3 {\tiny $\pm3.3$} \\
        %                                     & nBERTs &   74.2 {\tiny $\pm4.2$} &  51.7 {\tiny $\pm2.4$} &   67.8 {\tiny $\pm3.3$} &  76.9 {\tiny $\pm2.5$} &  67.7 {\tiny $\pm3.1$} \\
        % \midrule
        \multirow{2}{*}{\shortstack{\ours (Att.Q+LN) \\ ($6.84\%$)}} & \textbf{2.37e14} & Acc. &  \textbf{74.86} {\tiny $\pm5.27$}  &  55.81 {\tiny $\pm3.12$}  &   \textbf{66.99} {\tiny $\pm4.4$}  &  82.62 {\tiny $\pm2.73$}  &  \textbf{70.07} {\tiny $\pm3.88$}  \\
              & & nBERTs &  \textbf{69.02} {\tiny $\pm4.83$}  &  45.88 {\tiny $\pm3.72$}  &   \textbf{64.29} {\tiny $\pm4.7$}  &  75.63 {\tiny $\pm2.51$}  &   \textbf{63.7} {\tiny $\pm3.94$}  \\
        % \midrule
        % \multirow{2}{*}{LoRA} & Acc. &    67.7 {\tiny $\pm3.4$} $\triangledown$ &  39.84 {\tiny $\pm4.05$} $\triangledown$ &  63.32 {\tiny $\pm3.37$} $\triangledown$ &                 \textbf{84.12} {\tiny $\pm1.82$} $\triangledown$ &  63.74 {\tiny $\pm3.16$}  \\
        %       & nBERTs &  61.28 {\tiny $\pm3.12$} $\triangledown$ &   1.67 {\tiny $\pm1.57$} $\triangledown$ &    60.9 {\tiny $\pm3.3$} $\triangledown$ &                 \textbf{76.38} {\tiny $\pm1.66$} $\triangledown$ &  50.06 {\tiny $\pm2.41$}  \\
        \midrule
        \multirow{2}{*}{\shortstack{AdaLoRA \\ ($4.48\%$)}} & 2.87e14 & Acc. & 19.43  {\tiny $\pm1.47$} $\triangledown$ & 59.40 {\tiny $\pm2.28$} $\triangledown$ & 0.18 {\tiny $\pm0.20$} $\triangledown$ & \textbf{86.66} {\tiny $\pm1.79$} $\triangledown$ & 41.42 {\tiny $\pm1.44$} \\
            & & nBERTs & 16.26 {\tiny $\pm1.23$} $\triangledown$ & 48.30 {\tiny $\pm1.85$}  $\triangledown$ & 0.15 {\tiny $\pm0.16$} $\triangledown$ & 72.19 {\tiny $\pm1.49$} $\triangledown$ & 34.23 {\tiny $\pm1.18$} \\
        \midrule
        \multirow{2}{*}{\shortstack{AdaLoRA \\ ($1.15\%$)}} & 1.48e15 & Acc. & 69.66  {\tiny $\pm3.47$} $\triangledown$ & 46.60 {\tiny $\pm4.02$} $\triangledown$ & 61.80 {\tiny $\pm2.74$} $\triangledown$ & 84.50 {\tiny $\pm1.95$} $\triangledown$ & 65.64 {\tiny $\pm3.05$} \\
            & & nBERTs & 64.06 {\tiny $\pm3.19$} $\triangledown$ & 41.22 {\tiny $\pm3.65$}  $\triangledown$ & 58.91 {\tiny $\pm2.86$} $\triangledown$ & \textbf{77.43} {\tiny $\pm1.79$} $\triangledown$ & 60.41 {\tiny $\pm2.87$} \\
        \midrule
        \multirow{2}{*}{\shortstack{LoRA (Att.QV, Rank=128) \\ ($4.86\%$)}} & 2.88e14 & Acc. & 67.77 {\tiny $\pm3.73$} $\triangledown$ &  43.51 {\tiny $\pm3.57$} $\triangledown$ &  63.57 {\tiny $\pm3.16$} $\triangledown$ &  84.26 {\tiny $\pm1.92$} $\triangledown$ &   64.78 {\tiny $\pm3.1$}  \\
         &    & nBERTs &  61.36 {\tiny $\pm3.41$} $\triangledown$ &   0.33 {\tiny $\pm0.41$} $\triangledown$ &  61.06 {\tiny $\pm3.29$} $\triangledown$ &  76.49 {\tiny $\pm1.75$} $\triangledown$ &  49.81 {\tiny $\pm2.22$}  \\
        \midrule
        % \multirow{2}{*}{\shortstack{LoRA (Att.KQVO + FFN) \\ ($0.58\%$)}} & 2.75e14 & Acc. & 69.46 {\tiny $\pm4.16$} $\triangledown$ & 48.64 {\tiny $\pm4.05$} $\triangledown$ & 64.05 {\tiny $\pm3.59$} $\triangledown$ & 83.98 {\tiny $\pm2.18$} $\triangledown$ & 66.53 {\tiny $\pm3.50$} \\
        %     & & nBERTs & 63.98 {\tiny $\pm3.83$} $\triangledown$ & 42.98 {\tiny $\pm3.69$} $\triangledown$ & 61.37 {\tiny $\pm3.69$} $\triangledown$ & 76.96 {\tiny $\pm2.01$} $\triangledown$ & 61.32 {\tiny $\pm3.31$} \\
        % \midrule
        \multirow{2}{*}{\shortstack{LoRA (Att.KQVO, Rank=4) \\ ($0.32\%$)}} & 2.75e14 & Acc. & 68.96 {\tiny $\pm3.68$} $\triangledown$ & 39.04 {\tiny $\pm4.06$} $\triangledown$ & 62.66 {\tiny $\pm3.46$} $\triangledown$ & 84.05 {\tiny $\pm1.81$}$\triangledown$& 63.68 {\tiny $\pm3.25$} \\
            & & nBERTs & 63.48 {\tiny $\pm3.39$} $\triangledown$ & 33.52 {\tiny $\pm3.76$} $\triangledown$ & 59.80 {\tiny $\pm3.66$} $\triangledown$ & 77.04 {\tiny $\pm1.66$} $\triangledown$ & 58.46 {\tiny $\pm3.12$} \\
        \midrule
        \multirow{2}{*}{\shortstack{(IA)$^3$ \\ ($0.07\%$)}} & 2.74e14 & Acc. & 58.53 {\tiny $\pm2.32$} $\triangledown$ & \textbf{59.14} {\tiny $\pm2.36$} $\triangledown$ & 48.08 {\tiny $\pm0.81$} $\triangledown$ & 86.64 {\tiny $\pm1.85$} $\triangledown$ & 63.10 {\tiny $\pm1.84$} \\
            & & nBERTs & 53.87 {\tiny $\pm2.15$} $\triangledown$ & \textbf{48.08} {\tiny $\pm1.92$} $\triangledown$ & 48.06 {\tiny $\pm0.80$} $\triangledown$ &  72.18 {\tiny $\pm1.54$} $\triangledown$ & 55.55 {\tiny $\pm1.60$} \\
        % \midrule
        % \multirow{2}{*}{Prefix Tuning+LoRA} & Acc. & 71.31 {\tiny $\pm3.75$} & 47.71 {\tiny $\pm5.21$} & 61.68 {\tiny $\pm3.45$} & 84.67 {\tiny $\pm1.69$} & 66.34 {\tiny $\pm3.53$} \\
        %     & nBERTs & 65.66 {\tiny $\pm3.49$} & 42.48 {\tiny $\pm4.69$} & 59.16 {\tiny $\pm3.51$} & \textbf{77.56} {\tiny $\pm1.57$}  &  61.22 {\tiny $\pm3.32$}  \\
        % \midrule
        % \multirow{2}{*}{Prefix-Tuning} & Acc. &   1.79 {\tiny $\pm1.06$} $\triangledown$ &                 57.05 {\tiny $\pm2.33$}  &   0.02 {\tiny $\pm0.08$} $\triangledown$ &                 84.97 {\tiny $\pm3.44$}  &  35.96 {\tiny $\pm1.73$}  \\
        %       & nBERTs &     0.0 {\tiny $\pm0.0$} $\triangledown$ &     0.0 {\tiny $\pm0.0$} $\triangledown$ &   0.02 {\tiny $\pm0.07$} $\triangledown$ &     0.0 {\tiny $\pm0.0$} $\triangledown$ &    0.0 {\tiny $\pm0.02$}  \\
        \bottomrule
    \end{tabular}
    }
    \caption{Performance comparison between \ours and other PEFT strategies. We report the average and the standard deviation over the 60 few-shot train-validation splits for the \textbf{accuracy} metric and the normalized BERTScore~(\textbf{nBERTs}). We show in \textbf{bold} the setting with the highest metric for each dataset. Significance testing was assessed via mean t-test in comparison with \ours : $\triangledown$~represents a p-value lower than $10^{-2}$.}
    \label{tab:sparsefit_vs_peft}
\end{table*}

% Downstream Performance
\paragraph{Task Performance}%\label{sec:results_sparse_downstream}

We present in \cref{tab:sparse_summary_all} the accuracy performance for selected \ours settings for \texttt{T5-large}.
As can be observed in \cref{tab:sparse_summary_all}, some \ours configurations with very few fine-tuned parameters can produce significantly better results than the baseline~(\ie, full fine-tuning). 
For instance, fine-tuning the \emph{Normalization Layer}~(\emph{LayerNorm})~($0.02\%$ of the model's parameters) achieves better task performance than the baseline for two out of four datasets (e-SNLI and ECQA).
Furthermore, we consistently see that if two \ours configurations achieve good generalization results, combining them by jointly fine-tuning both components produces significantly better results than each configuration in isolation.
We show in \cref{fig:all_t5large_accuracy} the spread of the task performance for \ours configurations. 
Results for \texttt{T5-base} and \texttt{T5-3b} are shown in \cref{ap:sparse_full_results}.
%
% We include in \cref{ap:sparse_full_results} the task performance for all model sizes. 
%
We found that the task accuracy is consistently higher for the largest LMs for all datasets, but the gap between \texttt{T5-large} and \texttt{T5-3b} is small ($<$7\%) compared with the increase in trainable parameters (${\sim}$5x). 
% 

% Explanations Quality
% \vspace{-1em}
\paragraph{NLE Quality}%\label{sec:results_sparse_explanations}

Recall that the LM is fine-tuned to conditionally generate a text in the form of \textit{``[label] because [explanation]''}. 
\cref{fig:all_t5large_explanations} shows the normalized BERTScore for selected \ours settings as a proxy to evaluate how good the NLEs generated after the explanation token (\ie \emph{``because''}) is. For all the box plots, the $x$-axis shows the \ours configurations, with the percentage of fine-tuned parameters between brackets.
Overall, it can be observed that \ours settings with few trainable parameters (<$10\%$), such as the \textit{Self-attention Query}~(\emph{Att.Q}), \textit{LM Head + Attention Query}~(\emph{Att.Q+LMhead}), and \textit{Layer Norm + Attention Query}~(\emph{Att.Q+LN}), are competitive against the baseline for all datasets. 
Moreover, we can see that the best quality of NLEs is achieved for \ours combinations of two or more types of components (\eg, \emph{Att.Q}). 
Remarkably, fine-tuning the decoder block (${\sim}54\%$ params) achieves better performance than the entire fine-tuning for two out of four datasets (e-SNLI and ECQA).
The performance gap between most of the \ours configurations and the baseline does not exceed $15\%$ for all the datasets, even for very sparse fine-tuning strategies.
Unexpectedly, many \ours configurations with high task accuracy (\eg, \textit{LayerNorm}) have a normalized BERTScore close or equal to zero. This happens because either the conditional generation ends the sentence after the generated label token or the explanation token (\ie, \textit{``because''}) is not successfully generated. We investigate more about this behavior in  \cref{sec:discussion}.
We summarize our results on NLEs quality for \texttt{T5-large} in \cref{tab:sparse_summary_all}.
Results for other T5 model sizes (\ie \texttt{T5-base}, \texttt{T5-large} and \texttt{T5-3b}) are shown in \cref{ap:sparse_task_full_results}.
We found that the normalized BERTScore consistently increases with the size of the LM. Remarkably, the best \ours configurations for \texttt{T5-large} also achieve the best performance when fine-tuning \texttt{T5-base}, but they are slightly different for \texttt{T5-3b}.

% 

%%%%%%%%%%%%%%%%%%%%%%%%%%%%%%%%%%%%%%%%%%%%
%%%%%%%%%%%%%%%%%%%%%%%%%%%%%%%%%%%%%%%%%%%%
% Table for comparison with Parameter-Efficient Fine-Tuning (PEFT) strategies
\begin{table*}[t]
    \centering
    \resizebox{\textwidth}{!}{
    
    \begin{tabular}{>{\centering\arraybackslash}p{7cm} c c c c c c}
    \toprule
        %\textbf{Dataset} & \multicolumn{4}{c| }{Dataset} & \textbf{Avg} \\ \hline \hline
        % \midrule
          & & \bf ComVE & \bf ECQA & \bf SBIC & \bf e-SNLI &  \textbf{Avg} \\ %\hline
        \midrule
        %         \multirow{2}{*}{Baseline} & Acc. & 80.5 {\tiny $\pm4.5$} &  57.6 {\tiny $\pm2.6$} &   70.1 {\tiny $\pm3.4$} &   84.8 {\tiny $\pm2.6$} &  73.3 {\tiny $\pm3.3$} \\
        %                                     & nBERTs &   74.2 {\tiny $\pm4.2$} &  51.7 {\tiny $\pm2.4$} &   67.8 {\tiny $\pm3.3$} &  76.9 {\tiny $\pm2.5$} &  67.7 {\tiny $\pm3.1$} \\
        % \midrule
        \multirow{2}{*}{\shortstack{Baseline - Full Fine-tuning \\ ($100\%$)}} & Acc. &  63.71 {\tiny $\pm9.14$}  &  11.14 {\tiny $\pm3.14$}  & \textbf{63.86} {\tiny $\pm1.86$}  &  34.91 {\tiny $\pm0.43$}  & 43.41  {\tiny $\pm3.64$}  \\
              & nBERTs &  55.93 {\tiny $\pm9.16$}  &  9.46 {\tiny $\pm2.79$}  &   \textbf{57.42} {\tiny $\pm1.32$}  &  28.62 {\tiny $\pm0.84$}  &   37.86 {\tiny $\pm3.53$}  \\
        \midrule
        \multirow{2}{*}{\shortstack{\ours (Att.Q+LN) \\ ($7.97\%$)}}  & Acc. &  \textbf{68.03} {\tiny $\pm8.24$}  &  \textbf{24.53} {\tiny $\pm3.34$}  &   57.90 {\tiny $\pm1.70$}  &  \textbf{40.10} {\tiny $\pm4.03$}  &  \textbf{47.64} {\tiny $\pm4.33$}  \\
              & nBERTs &  \textbf{58.67 }{\tiny $\pm7.20$}  &  \textbf{20.60} {\tiny $\pm2.83$}  &   50.41 {\tiny $\pm2.72$}  &  \textbf{34.32} {\tiny $\pm3.50$}  &  \textbf{41.00}  {\tiny $\pm4.06$}  \\
        \midrule
        \multirow{2}{*}{\shortstack{AdaLoRA\\ ($0.30\%$)}} & Acc. &  64.23 {\tiny $\pm2.86$}  &  13.04 {\tiny $\pm2.09$}  &   57.29 {\tiny $\pm1.86$}  &  38.15 {\tiny $\pm4.22$}  &  43.18 {\tiny $\pm2.76$}  \\
              & nBERTs &  56.16 {\tiny $\pm2.77$}  &  11.13 {\tiny $\pm1.79$}  &   50.63 {\tiny $\pm0.33$}  &  33.48 {\tiny $\pm3.71$}  &  37.85  {\tiny $\pm2.15$}  \\
        
        \bottomrule
    \end{tabular}
    }
    \caption{Performance comparison between \ours and other PEFT strategies for \textbf{Llama 2-7B}. We report the average and the standard deviation over the 60 few-shot splits for the \textbf{accuracy} metric and the normalized BERTScore~(\textbf{nBERTs}). We show in \textbf{bold} the setting with the highest metric for each dataset. Significance testing was assessed via mean t-test in comparison with \ours : $\triangledown$~represents a p-value lower than $10^{-2}$.}
    \label{tab:llama_sparsefit_vs_peft}
\end{table*}

%%%%%%%%%%%%%%%%%%%%%%%%%%%%%%%%%%%%%%%%%%%%
%%%%%%%%%%%%%%%%%%%%%%%%%%%%%%%%%%%%%%%%%%%%%%%%%%%%%%%%%%%%%%%%%%%%%%%%%%%%%%%%%%%%%%%%
%%%%%%%%%%%%%%%%%%%%%%%%%%%%%%%%%%%%%%%%%%%%%%%%%%%%%%%%%%%%%%%%%%%%%%%%%%%%%%%%%%%%%%%%
%%%%%%%%%%%%%%%%%%%%%%%%%%%%%%%%%%%%%%%%%%%%

% Comparison against other PEFT strategies
% \vspace{-1em}
\paragraph{Other PEFT Baselines}\label{sec:results_other_peft}
% To compare \ours with other PEFT baselines, we also implemented LoRA~\cite{hu2022lora} and Prefix-Tuning~\cite{li2021prefixtuning} for NLEs. \cref{tab:sparsefit_vs_peft} shows the downstream performance and the NLEs quality for different PEFT strategies on \texttt{T5-large}. It can be seen that, on average, \ours outperforms the other two PEFT strategies. The best PEFT approach is LoRA with a performance roughly 10\% lower than \ours. Furthermore, the quality of NLEs is considerably better for \ours for 3 out of 4 datasets.

To compare \ours with other PEFT baselines, we also evaluated LoRA~\cite{hu2022lora}, AdaLoRA~\cite{adalora} and (IA)$^3$~\cite{ia3}
%
% and Prefix-tuning~\cite{li2021prefixtuning}
%
for NLEs.
%
% For LoRA~\cite{hu2022lora}, we performed a grid search over the rank of the injected matrices and found that the best results are obtained after injecting and updating matrices of \emph{rank=4}.
%
% The hybrid method combining LoRA~\cite{hu2022lora} with Prefix-Tuning~\cite{li2021prefixtuning} uses this best LoRA setting and a prefix length of 10.
%
\cref{tab:sparsefit_vs_peft} shows the downstream performance and the NLEs quality for different PEFT strategies on \texttt{T5-large}. It can be seen that, on average, \ours outperforms the other PEFT strategies. 
%
% AdaLoRA, LoRA, (IA)$^3$ and Prefix-Tuning+LoRA respectively add and update 8.65, 2.36, 0.54 and 2.85 million parameters. 
%
While these PEFT methods tune less than 20\% of the 50.45 million parameters updated by \ours, the quality of NLEs is considerably better for \ours for two out of four datasets. 
%
% Notice that LoRA performance is low for the ECQA dataset in terms of NLEs quality. We hypothesize that this behavior is connected with the fact that is more difficult to generate the explanation token for tasks with no common classes. 
%
Notice that in \cref{tab:sparsefit_vs_peft} \ours has the lowest FLOPS. We hyphotesise that this happens since \ours neither introduces additional model parameters nor increases the model's architectural complexity.
In \cref{tab:lora_higher_rank} in \cref{ap:further_peft_exploration}, we show further results for LoRA trained on a bigger range of the number of parameters.

%%%%%%%%%%%%%%%%%%%%%%%%%%%%%%%%%%%%%5

% Human evaluation summary
\begin{table}[t]
    \centering
    \resizebox{\columnwidth}{!}{
    
    \begin{tabular}{c c c c c c}
        \toprule
        \multicolumn{6}{c}{\textbf{Human Evaluation}} \\
        \midrule
        %\textbf{\ours} & \multicolumn{4}{c}{\textbf{Dataset}} & \textbf{Avg} \\
        %\midrule
          
         &                  e-SNLI &                    ECQA &                    SBIC &                   ComVE & \bf Avg \\
        \midrule
        \makecell{Full \\ Fine-tuning}               &  29.63 {\tiny ($0.43$)} &  \textbf{41.92} {\tiny ($0.23$)} &   54.44 {\tiny $\pm0.7$} &  21.67 {\tiny ($0.22$)} &   36.91 \\ 
        \midrule
        % \makecell{\ours \\ Decoder}                &  22.22 {\tiny $\pm0.45$} &  \textbf{49.49} {\tiny $\pm0.31$} &   60.0 {\tiny $\pm0.58$} &  15.56 {\tiny $\pm0.21$} &   36.82 \\
        % \midrule
        % encoder                &  19.14 {\tiny ($0.24$)} &  43.43 {\tiny ($0.22$)} &  56.67 {\tiny ($0.94$)} &   28.33 {\tiny ($0.3$)} &   36.89 \\
        \makecell{\ours \\ Att.Q}        &   17.28 {\tiny $0.38$} &   35.35 {\tiny ($0.31$)} &   \textbf{61.11} {\tiny ($0.77$)} &   28.89 {\tiny ($0.35$)} &   35.66 \\
        \midrule
        \makecell{AdaLora \\ 1.15\% } &   23.33 {\tiny ($0.34$)} &   34.44 {\tiny ($0.26$)} &   \textbf{61.11} {\tiny ($0.69$)} &    23.34 {\tiny ($0.25$)} &   35.55 \\
        \midrule
        \makecell{\ours \\ Att.Q+LN} &   \textbf{38.27} {\tiny $(0.34)$} &   31.31 {\tiny $(0.26)$} &   58.89 {\tiny ($0.69$)} &    \textbf{40.00} {\tiny ($0.25$} &   \textbf{42.12} \\
        \bottomrule
    \end{tabular}
    
    }
    \caption{Average scores given by human annotators for the best performing \ours and other PEFT baselines of \texttt{T5-large}. 
    The best results are in \textbf{bold}. In brackets, we show the inter-annotator agreement score. 
    % LN and SelfQA stand for \emph{LayerNorm} and \emph{Self-attention Query}, respectively.
    }
    \label{tab:sparse_summary_human}
\end{table}

%%%%%%%%%%%%%%%%%%%%%%%%%%%%%%%%%%%%%%

% Large Language Models
%\vspace{-1em}
\paragraph{Larger Language Models}\label{sec:results_large_languages}

To evaluate the performance of \ours in both larger language models and different architectures we perform a set of experiments applying \ours to \texttt{Llama 2-7B}. Notice that \ours approach applies to any architecture (not only the \texttt{T5} encoder-decoder) since it focuses on identifying the optimal layer to fine-tune, regardless of the underlying model's structure. In this regard, we conducted experiments on a larger decoder-only model (i.e. \texttt{Llama 2-7B}). \cref{tab:llama_sparsefit_vs_peft} shows the average predictive accuracy and the NLEs quality for the the best \ours strategy (Att.Q+LN) and the best performing PEFT baseline (AdaLora) for \texttt{Llama 2-7B}. Overall, it can be observed that \ours outperforms the other PEFT strategy for all datasets. Particularly, the best \ours have on average roughly $5$\% better NLE quality than the other PEFT.

% Human Evaluation
%\vspace{-1em}
\paragraph{Human Evaluation}\label{sec:results_sparse_human}

\begin{figure}[t]
    \centering
    \includegraphics[width=\columnwidth]{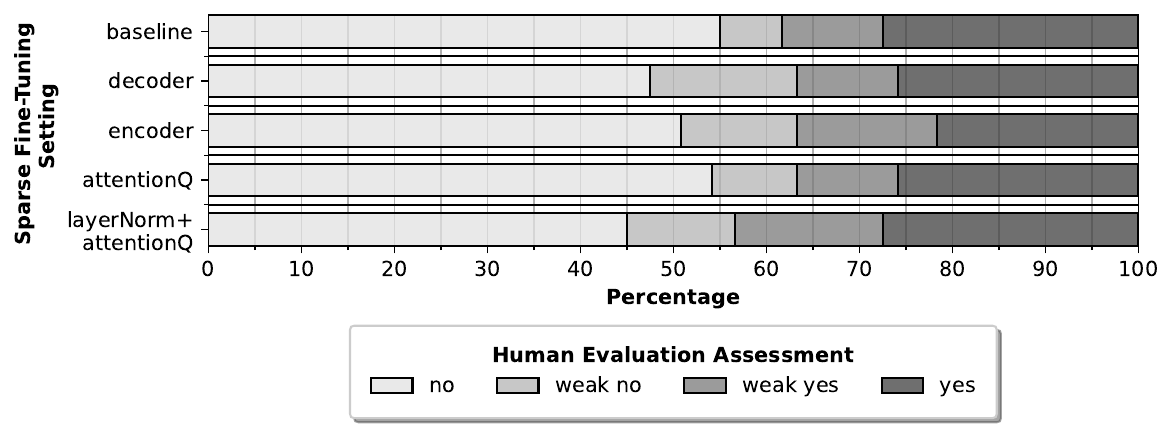}
    \caption{Illustration of plausibility score given by human annotators to the quality of the NLEs generated by different \ours configurations. The annotators were asked to answer the question: \textit{``Does the explanation justify the answer?}}%
    \label{fig:human_plausibility_all}
    % \vspace{-1em}
\end{figure}

%The goal of the human evaluation protocol is to evaluate how plausible are the predicted NLEs from the perspective of a human annotator for the best \ours settings. Furthermore, we aim to study to what extent the BERTScore (and the automatic metrics found in Section~\ref{ap:sparse_full_results}) are correlated with human judgement.
%
We show in~\cref{tab:sparse_summary_human} the distributions of the scores given by the human annotators for the quality of the generated NLEs for the best \ours strategies and the best performing PEFT baseline (AdaLora).
We compute the inter-annotator agreement score using the \emph{Cohen' $\kappa$} metric~\citep{cohen1960kappa}.
Overall, we found that the quality of the NLEs generated after applying \ours is much higher than those of the baseline and AdaLoRA for 2 out of 4 tasks. 
For the other two tasks, AdaLoRA produces better NLEs by a very smaller margin.
On average, the NLEs of \ours are roughly 8\% better than NLEs of AdaLoRA and 6\% better than full fine-tuning NLEs.
However, the human evaluation shows that the generated NLEs are often insufficient to explain the predictions accurately. 
We show in~\cref{fig:human_plausibility_all} the distributions of the plausability score given by the human annotators for the quality of the generated NLEs for the best \ours strategies.
It can be observed that roughly half of the NLEs do not justify the answer, no matter what fine-tuning strategy is used. Similarly, the proportion of NLEs that fully justify the prediction is close to $25\%$ regardless of the \ours setting.
We detail the shortcomings and limitations of generated NLEs in \cref{sec:discussion}. 
%
% Finally, \cref{tab:sparse_summary_human} shows the aggregated performance after remapping the plausibility scores to numerical values for different \ours configurations when applied to a larger language model (\ie \texttt{Llama 2-7B}).
Finally, we show in \cref{tab:llama_summary_human} the human evaluation results for the best \ours configuration and other PEFT baselines when applied to \texttt{Llama2-7B}. It can be seen that on average the NLEs of \ours are roughly 6\% better than NLEs of AdaLoRA and 21\% better than full fine-tuning NLEs of \texttt{Llama2-7B}.
%

%
% We compute the inter-annotator agreement score using the \emph{Cohen' $\kappa$} metric~\citep{cohen1960kappa}, which is suitable for our setting since we only have two annotators evaluating the same examples. 
% 

% Llama Human evaluation summary
\begin{table}[t]
    \centering
    \resizebox{\columnwidth}{!}{
    
    \begin{tabular}{p{3cm} c c c c c}
        \toprule
        \multicolumn{6}{c}{\textbf{Human Evaluation}} \\
        \midrule
          
         &                  e-SNLI &                    ECQA &                    SBIC &                   ComVE & \bf Avg \\
        \midrule
        \makecell{Full \\ Fine-tuning} &  17.78 &  38.89  &   47.78  &  40.00  &  36.11  \\ 
        \midrule
        \makecell{\ours \\  Att.Q+LN}  &  41.11  &  \textbf{73.33}  &  68.89   &  \textbf{70.00}   & \textbf{63.33 }  \\
        \midrule
        \makecell{AdaLora } & \textbf{50.00}  &  52.22  &  \textbf{71.11}  & 55.56   &  57.22  \\
        
        \bottomrule
    \end{tabular}
    
    }
    \caption{Average scores given by human annotators for the best performing \ours and other PEFT baselines of \texttt{Llama-2-7B}. 
    The best results are in \textbf{bold}. 
    % LN and SelfQA stand for \emph{LayerNorm} and \emph{Self-attention Query}, respectively.
    }
    \label{tab:llama_summary_human}
\end{table}

%%%%%%%%%%%%%%%%%%%%%%%%%%%%%%%%%%%%%%%%%%%%%%%%%%%%%
% ===================================================
% Discussion
% ===================================================
%%%%%%%%%%%%%%%%%%%%%%%%%%%%%%%%%%%%%%%%%%%%%%%%%%%%%
\subsection{Discussion}\label{sec:discussion}

% Examples of predicted explanations
%\vspace{-1em}
\paragraph{Analysis of the Generated NLEs} %\label{sec:discussion_explanation_examples}
In \cref{fig:examples_nle_all}, we show a collection of examples of the generated NLEs by the baseline and the best performing \ours configurations.
As in previous works~\citep{camburu2018snli,kayser2021vil,marasovic-etal-2022-shot}, we only show examples where the label was correctly predicted by the model since we do not expect a model that predicted a wrong label to generate a correct explanation. 
We present in \cref{ap:more_examples_explanations} a more extensive list of generated NLEs with \ours.
%

% Explanations Shortcomings
% \subsubsection{NLE Shortcomings}\label{sec:discussion_explanation_shortcomings}
%\vspace{-1em}
\paragraph{NLE Shortcomings}

\begin{figure}[t]
    \centering
    \includegraphics[width=0.5\textwidth]{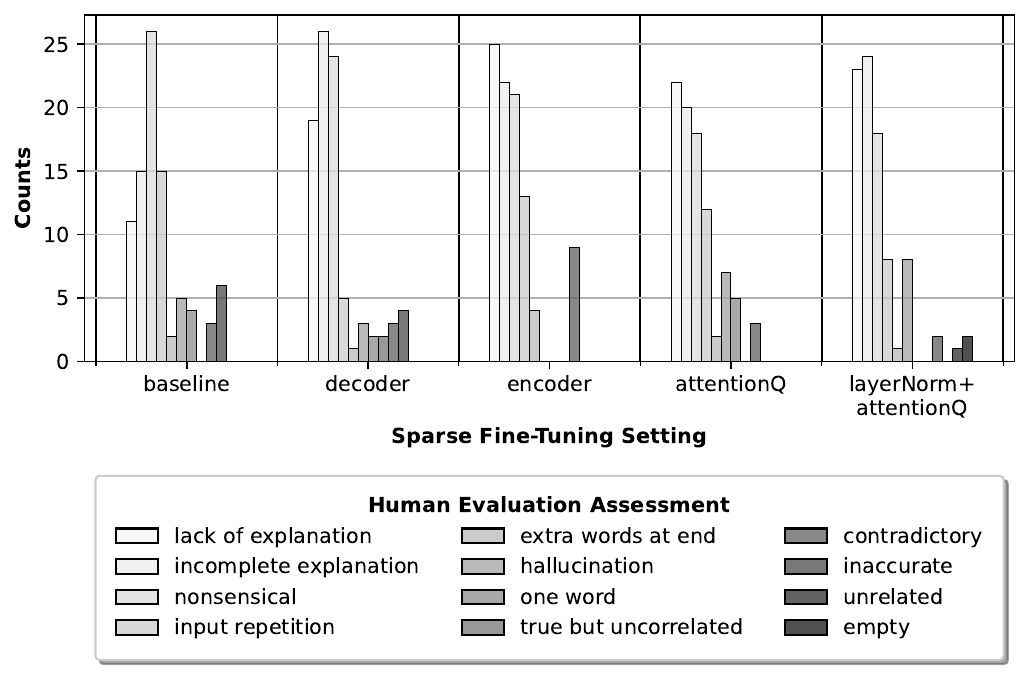}
    \caption{Histogram of the shortcomings of the generated NLEs for the baseline and the performing \ours configurations aggregated for all the datasets.}
    \label{fig:human_plausibility_reason_all}
    % \vspace{-1em}
\end{figure}

\cref{fig:human_plausibility_reason_all} depicts the histogram of frequencies of the shortcomings for the baseline and the best-performing \ours strategies.
It can be observed that the most common shortcomings are \textit{Lack of explanation}, \textit{Nonsensical}, and \textit{Incomplete explanation}. 
%
%Remarkably, there is not a common ranking of the frequency of shortcomings over the different \ours settings.
%
% Remarkably, different \ours settings produce different distributions over the shortcomings.
%
For the best \ours configuration~(\ie \emph{Att.Q+LN}), the \textit{Incomplete explanation} is the reason with the most occurrences. 
% , followed by \textit{Lack of explanation}.
%
We show a breakdown of the shortcomings for each dataset in \cref{ap:shortcomings_per_dataset}.

% Plot Human Plausibility - All

% Examples all datasets
\begin{figure}[t!]
        \begin{tcolorbox}[colback=green!5!white, colframe=black,
        left=1mm,top=1mm,right=1mm,bottom=1mm, fonttitle=\small]
        \resizebox{\columnwidth}{!}{
        \begin{tabular}{ p{7.5em} | p{23.5em} } 
            \textbf{Premise}  & 2 woman walking down the street next to graffiti \\ 
            \textbf{Hypothesis} & The 2 woman are running \\
            \textbf{Label} & Contradiction \\
            \textbf{Explanation} & You can't be running, if you are walking. Walking is different from running. \\
            &  \\
            &  \textbf{Generated Explanation}  \\
            \textbf{Baseline} & walking and running are different actions \\ 
            \textbf{Decoder}  & walking and running are different verbs \\ 
            % \textbf{Encoder} & walking does not imply running \\
            \textbf{Att.Q} & the 2 woman are walking not running \\
            \textbf{Att.Q + Head} & walking and running are not the same \\
            \textbf{Att.Q + LN} & walking is not running \\
            % \textbf{Attention KQV} & walking and running are not the same
        \end{tabular}
        }
        \end{tcolorbox}
        \begin{tcolorbox}[colback=blue!5!white, colframe=black,
        left=1mm,top=1mm,right=1mm,bottom=1mm, fonttitle=\small]
        \centering
        \resizebox{\columnwidth}{!}{
        \begin{tabular}{ p{7.5em} | p{23.5em} } 
            \textbf{Question} & The teacher doesn't tolerate noise during a test in their what? \\ 
            \textbf{Answer Choices} & A) Theatre B) Bowling C) Alley D) Factory E)  Classroom \\
            \textbf{Answer} & Classroom \\
            \textbf{Explanation} & Test is done in classroom. A teacher when in classroom does not tolerate noise.\\
            &  \\
            &  \textbf{Generated Explanation}  \\
            \textbf{Baseline} & noise is harmful for students. classroom is a place where students can take test. \\ 
            \textbf{Decoder}  &  classroom is a place where students are studying. there is no noise in classroom. \\ 
            % \textbf{Encoder} & classroom is a place where teacher doesn't tolerate noise during a test. \\
            \textbf{Att.Q} & classroom is a place where people can learn. noise is a problem in classroom.\\         
            \textbf{Att.Q + Head} & noise is harmful to students \\
            \textbf{Att.Q + LN} & classroom is a place where people can learn. noise is a distracting thing. \\
            % \textbf{Attention KQV} & noise is a threat to the students' concentration. \\
        \end{tabular}
        }
        \end{tcolorbox}
        \begin{tcolorbox}[colback=red!5!white, colframe=black,
        left=1mm,top=1mm,right=1mm,bottom=1mm, fonttitle=\small]
        \centering
        \resizebox{\columnwidth}{!}{
        \begin{tabular}{ p{7.5em} | p{23.5em} } 
            \textbf{Post}  & @cmencarini: I dont trust girls ; cant trust these hoes \\ 
            \textbf{Label} & Offensive \\
            \textbf{Explanation} &  This post implies that women are not honest. This post implies that women are not trustworthy. This post implies that women are all women are hoes. \\
            &  \\
            &  \textbf{Generated Explanation}  \\
            \textbf{Baseline} & this post implies that women aren't important \\ 
            \textbf{Decoder}  & this post implies that women are prostitutes \\ 
            % \textbf{Encoder} & this post implies that women are not trustworthy \\
            \textbf{Att.Q} & this post implies that women aren't trustworthy \\
            \textbf{Att.Q + Head} & this post implies that women aren't trustworthy \\
            \textbf{Att.Q + LN} & this post implies that women are dirty \\
            % \textbf{Attention KQV} & this post implies that women aren't worth trusting \\
        \end{tabular} 
        }
        \end{tcolorbox}
        \begin{tcolorbox}[colback=yellow!5!white, colframe=black,
        left=1mm,top=1mm,right=1mm,bottom=1mm, fonttitle=\small]
        \centering
        \resizebox{\columnwidth}{!}{
        \begin{tabular}{ p{7.5em} | p{23.5em} } 
            \textbf{Choice 1} & A Latino can communicate with someone from Spain \\ 
            \textbf{Choice 2} & French and Latinos can communicate easy \\
            \textbf{Label} & Choice 2 \\ 
            \textbf{Explanation} & French can't communicate with Latinos because they don't speak the same language. French and Latinos belongs to two different countries with two different languages. \\ 
            &  \\
            &  \textbf{Generated Explanation}  \\
            \textbf{Baseline} & french and latinos are two different languages \\ 
            \textbf{Decoder}  & french and latinos are two different people and don't speak the same language \\ 
            % \textbf{Encoder} & french is not a language that latinos can use \\
            \textbf{Att.Q} & french is not a common language in latinamerica \\
            \textbf{Att.Q + Head} & french and latinos cannot communicate easily. \\
            \textbf{Att.Q + LN} & french and latinos cannot communicate easily \\
            % \textbf{Attention KQV} & french language isn't used in latinamerica \\
        \end{tabular}
        }
        \end{tcolorbox}
    %\vspace{-1em}
    \caption{Examples of generated NLEs for e-SNLI~{\small(Green)}, ECQA~{\small(Blue)}, SBIC~{\small(Red)}, and ComVE~{\small(Yellow)}.}
    \label{fig:examples_nle_all}
    \vspace{-1.2em}
\end{figure}

% Inter-annotator agreement
% \subsubsection{Inter-Annotator Agreement}\label{sec:discussion_annotator_agreement}
\paragraph{Inter-Annotator Agreement}
As shown in \cref{tab:sparse_summary_human}, the agreement between annotators is moderately low for the set of evaluated NLEs.
More precisely, the annotators gave different scores to 181 out of 600 NLEs.
The dataset with the most significant difference is ECQA, with 63 differences, while the SBIC dataset is the most uniform, with 17 differences.
The variation between annotators can result from three potential perceptual reasons ~\cite{bourke2014positionality,nino2009machine}.
%
% The first reason is the difference in interpretation of scores within the spectrum of possible answers (\ie \emph{No}, \emph{Weak No}, \emph{Weak Yes}, \emph{Yes}).
%
The first reason is the \emph{perceptual disagreement}, which states that annotators could not objectively identify the difference between two adjacent answers (\ie \emph{Weak Yes} vs \emph{Weak No}, or \emph{Yes} vs. \emph{Weak Yes}).
%
%This ambiguity leads to different perceptions about the completeness or failure of a particular explanation.
%
% The second reason is \emph{positionality disagreement}~\citep{bourke2014positionality}, which refers to the differences in the positions of the annotators due to their race, gender, and other socioeconomic identity factors that can alter the way they perceive the outcomes of the algorithm. This is particularly crucial for the SBIC dataset, as it contains offensive content.
%
The second reason is \emph{positionality disagreement}~\citep{bourke2014positionality}, which could alter how the annotators perceive the outcomes of the algorithm due to their race, gender, and other socioeconomic identity factors. This is particularly crucial for the SBIC dataset, as it contains offensive content.
The third reason is the \emph{expectation disagreement}, which may cause an annotator to be more strict on the characteristics that make an explanation complete and accurate.
%
% The third reason is related to the expectations about the quality of the generated NLEs. In this regard, an annotator may be more strict on the characteristics that make an explanation complete and accurate.
%
% \cref{fig:examples_expectation_disagreement} shows an example of expectation disagreement. See \cref{ap:examples_inter_agreement} for more examples.
%
An extensive collection of examples of perceptual disagreement, positionality disagreement, and expectation disagreement samples are in  \cref{ap:examples_inter_agreement}.

% Explanations Shortcomings
% \subsubsection{Generation of Empty NLEs}\label{sec:discussion_empty_explanations}
\paragraph{Generation of Empty NLEs}
As mentioned earlier, some \ours configurations~(\eg \emph{LayerNorm}) have high task performance but generate empty NLEs most of the time, particularly for the e-SNLI and ECQA datasets.
One possible explanation for the discrepancy between the high task accuracy and the low NLE quality is that generating NLEs is an intrinsically more complex problem than solving the downstream tasks, where the former may require fine-tuning more significant portions of the model parameters.
Another explanation can be found by analyzing the pre-training tasks of the PLM and observing that, in the pre-training stage, T5 was trained on the MNLI dataset~\cite{williams2018mnli}, which is composed of NLI instances without NLEs.
The T5 weights were then pre-trained on MNLI by casting the NLI task as a sequence transduction problem, where the input is a hypothesis-premise pair, and the output is the label. When only a small subset of parameters is updated (\eg, \emph{LayerNorm}($0.02\%$)), the model elicits its original behavior and predicts the label %with high accuracy 
without generating the NLE.
Similar reasoning may be concluded for ECQA since \textsc{UnifiedQA} was pre-trained on CommonsenseQA~\citep{talmor-etal-2019-commonsenseqa}, which is composed of samples with only the answer for the multiple-choice question.

%%%%%%%%%%%%%%%%%%%%%%%%%%%%%%%%%%%%%%%%%%%%%%%%%%%%%
% ===================================================
% Conclusion
% ===================================================
%%%%%%%%%%%%%%%%%%%%%%%%%%%%%%%%%%%%%%%%%%%%%%%%%%%%%
%\section{Conclusions and Future Work}
\section{Summary}
%In this work, w
We introduced \ours, a strategy that combines sparse fine-tuning with prompt-based learning to train NLE models in a few-shot setup. 
\ours shows consistently competitive performance while only updating a minimal subset of parameters (\ie the \emph{Self-attention Query + Layer Normalization}, having ${\sim}6.8\%$ of the model parameters). 
%
%Concerning the best \ours configuration, 
%Moreover, we found that fine-tuning the \emph{Self-attention Query plus Layer Normalization} achieves the best trade-off between generalization performance and the percentage of fine-tuned parameters.
%
% The \emph{Decoder} and  the \emph{Self-attention Layer} %and the %\emph{Self-attention Query} 
% are also consistently among the three best performers. % for accuracy and NLE quality.
%
%However, we did not find a standard ranking across the evaluated datasets. 
%
We found that the sparse fine-tuning of \texttt{T5-large} consistently achieves better performance than fine-tuning \texttt{T5-base} and is slightly worse ($<5\%$) than \texttt{T5-3b}, no matter the \ours strategy. Moreover, the top three best performers for \texttt{T5-base} are achieved by the same set of \ours configurations found for \texttt{T5-large}.
Compared to other PEFT techniques, \ours produces better %results on 
average %in terms of 
predictive accuracy and NLE quality.
We aim for \ours to inspire the community to investigate sparse fine-tuning at different model components. %Future work may also build upon \ours by, \eg relying on soft prompts rather than hard prompts.

\section*{Limitations}

Although generating natural language explanations is a fervid research area, there is still no guarantee that such explanations accurately reflect how the model works internally~\cite{wiegreffe-etal-2021-measuring, DBLP:conf/acl/CamburuSMLB20}.
For example, the fact that the generated explanation seems reasonable does not mean that the model does not rely on protected attributes and spurious correlations in the training data to produce its predictions.
As such, we still recommend being careful to use self-explanatory models in production, as they can capture potentially harmful biases from the training data, even though these are not mentioned in the explanations.
%

% Todo: uncomment back for camera-ready
\section*{Acknowledgments}
We thank Andrea Sissa for helping with the human evaluation and for her insightful ideas on the inter-annotator agreement variations. 
Oana-Maria Camburu was supported by a Leverhulme Early Career Fellowship.
Pasquale Minervini was partially funded by the European Union’s Horizon 2020 research and innovation programme under grant agreement no. 875160, ELIAI (The Edinburgh Laboratory for Integrated Artificial Intelligence) EPSRC (grant no. EP/W002876/1), an industry grant from Cisco, and a donation from Accenture LLP.
%

%%%%%%%%%%%%%%%%%%%%%%%%%%%%%%%%%%%%%%%%%%%%%%%%%%%%%
% ===================================================
% Bibliography
% \bibliographystyle{acl_natbib_n}
\bibliography{anthology,custom}

\clearpage

%%%%%%%%%%%%%%%%%%%%%%%%%%%%%%%%%%%%%%%%%%%%%%%%%%%%%
% ===================================================
% Appendices
% ===================================================
%%%%%%%%%%%%%%%%%%%%%%%%%%%%%%%%%%%%%%%%%%%%%%%%%%%%%
\appendix

\section{\ours Graphical Representation}\label{ap:sparse_full_diagrams}

\begin{figure}[h]
    \centering
    \includegraphics[width=\columnwidth]{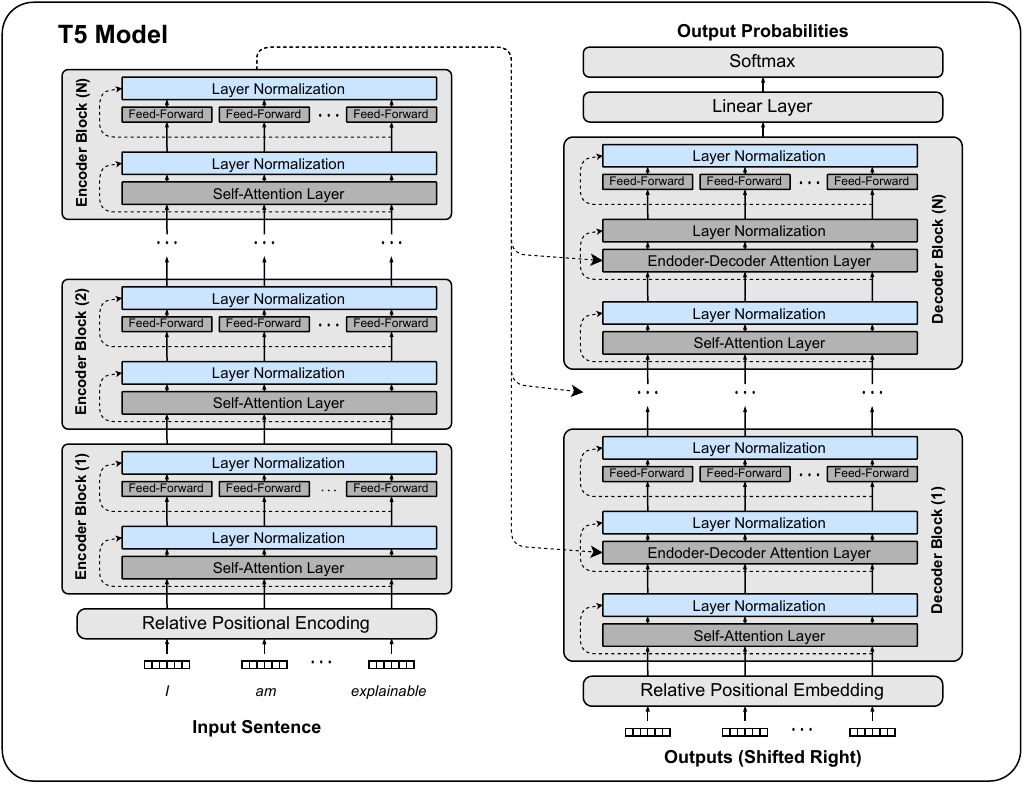}
    \caption{Illustration of the active trainable parameters in T5 when \ours is performed over the layer normalization. 
    }
    \label{fig:t5_sparse_norm}
\end{figure}

In this paper, we propose an efficient few-shot prompt-based training regime for models generating both predictions and NLEs on top of the T5 language model. To have a better understanding of the active trainable parameters in each \ours configuration, we illustrate in \cref{fig:t5_sparse_norm} a graphical representation of the T5 architecture with active parameters colored for the \emph{Layer Normalization} sparse fine-tuning. After freezing the rest of the model (gray-colored layers), the percentage of parameters that could potentially be updated in the \emph{Layer Normalization} is $0.02\%$ of the entire model. Considering that the \textsc{UnifiedQA} model's architecture is the same as the one in T5, the interpretation of active parameters holds for \textsc{UnifiedQA}.

% Layer Normalization illustration

%%%%%%%%%%%%%%%%%%%%%%%%%%%%%%%%%%%%%%%%%%%%%%%%%%%%%%%%
%%%%%%%%%%%%%%%%%%%%%%%%%%%%%%%%%%%%%%%%%%%%%%%%%%%%%%%%
%%%%%%%%%%%%%%%%%%%%%%%%%%%%%%%%%%%%%%%%%%%%%%%%%%%%%%%%

\section{Examples of Generated NLEs}\label{ap:more_examples_explanations}

This section shows a collection of examples of the generated NLEs by the baseline and the different sparse fine-tuning strategies considered in our approach. We show four examples for each dataset. Each example contains the generated NLE for the best performing \ours configurations. As in previous works \citep{camburu2018snli, kayser2021vil, marasovic-etal-2022-shot}, we only show examples where the label was correctly predicted by the model (since we do not expect a model that predicted a wrong label to generate a correct NLE). 
Regarding the protocol for choosing the examples shown in this section, we have done a manual inspection of several possible examples, and we have chosen the more informative ones to conclude the strengths and the weaknesses of the generated NLEs. Notice that, due to the few-shot splits protocol (60 different train-validation splits), a single example could be predicted more than once for a single setup (i.e., the sample is in more than one validation set). 
%Considering that we have chosen manually the examples shown in this section, for each sparse fine-tuning setup, only the best explanation among the multiple generated explanations for those samples is shown. Below, you can find the set of examples for all four datasets. 

% Examples e-SNLI
\begin{figure}[h]
        \begin{tcolorbox}[colback=green!5!white, colframe=black,
        left=1mm,top=1mm,right=1mm,bottom=1mm, fonttitle=\small]
        \resizebox{\columnwidth}{!}{
        \begin{tabular}{ p{8em} | p{23em} } 
            \textbf{Premise}  & Two dogs play together on the bed. \\ 
            \textbf{Hypothesis} & Dogs laying down on the floor, motionless. \\
            \textbf{Label} & Contradiction \\
        
            &  \\
            &  \textbf{Generated Explanation}  \\
            \textbf{Baseline} & dogs can't be motionless and play together on the bed. \\ 
            \textbf{Decoder}  & playing together is different than lying down, motionless. \\ 
            \textbf{Encoder} & dogs laying down on the floor are motionless. \\
            \textbf{Attention.Q} & two dogs playing together on the bed are not laying down, motionless. \\
            \textbf{LM Head + Att.Q} & two dogs playing together on the bed are not motionless. \\
            \textbf{LayNorm + Att.Q} & two dogs playing together on the bed are not laying down, motionless. \\
            \textbf{Self-Attention} & two dogs playing together on the bed are not motionless. \\
        \end{tabular}
        }
        \end{tcolorbox}
        \begin{tcolorbox}[colback=green!5!white, colframe=black,
        left=1mm,top=1mm,right=1mm,bottom=1mm, fonttitle=\small]
        \resizebox{\columnwidth}{!}{
        \begin{tabular}{ p{8em} | p{23em} } 
            \textbf{Premise}  &  Man walking barefoot on the beach on what seems to be a very cloudy day \\ 
            \textbf{Hypothesis} & It is going to rain soon.g \\
            \textbf{Label} & Neutral \\
        
            &  \\
            &  \textbf{Generated Explanation}  \\
            \textbf{Baseline} & it does not look like it is going to rain soon \\ 
            \textbf{Decoder}  & not all cloudy means it is raining \\ 
            \textbf{Encoder} & it does not mean it is going to rain soon \\
            \textbf{Attention.Q} & not all clouds are bad \\
            \textbf{LM Head + Att.Q} & not all cloudy days are going to rain soon \\
            \textbf{LayNorm + Att.Q} & it does not mean it is going to rain soon \\
            \textbf{Self-Attention} &  not all cloudy days are going to rain soon \\
        \end{tabular}
        }
        \end{tcolorbox}
        \begin{tcolorbox}[colback=green!5!white, colframe=black,
        left=1mm,top=1mm,right=1mm,bottom=1mm, fonttitle=\small]
        \resizebox{\columnwidth}{!}{
        \begin{tabular}{ p{8em} | p{23em} } 
            \textbf{Premise}  & Number 13 kicks a soccer ball towards the goal during children's soccer game \\ 
            \textbf{Hypothesis} & A player trying to score in a soccer game \\
            \textbf{Label} & Entailment \\
        
            &  \\
            &  \textbf{Generated Explanation}  \\
            \textbf{Baseline} & number 13 is a player \\ 
            \textbf{Decoder}  & number 13 is a player trying to score \\ 
            \textbf{Encoder} & number 13 is a player trying to score \\
            \textbf{Attention.Q} & number 13 is a player \\
            \textbf{LM Head + Att.Q} & number 13 is a player trying to score \\
            \textbf{LayNorm + Att.Q} & number 13 is a player \\
            \textbf{Self-Attention} &  kicking a soccer ball towards the goal is similar to trying to score \\
        \end{tabular}
        }
        \end{tcolorbox}
    %\vspace{-1em}
    \caption{Examples of generated NLEs for e-SNLI.}
    \label{fig:examples_nle_esnli}
\end{figure}

% Examples ECQA
\begin{figure}[H]
        \begin{tcolorbox}[colback=blue!5!white, colframe=black,
        left=1mm,top=1mm,right=1mm,bottom=1mm, fonttitle=\small]
        \centering
        \resizebox{\columnwidth}{!}{
        \begin{tabular}{ p{8em} | p{23em} } 
            \textbf{Question}  & What does a person who likes school also do? \\ 
            \textbf{Answer Choices} & A) Fever B) Play dumb C) Read text D) Catch cold E) Enjoy learning \\
            \textbf{Answer} & Enjoy learning \\
            &  \\
            &  \textbf{Generated Explanation}  \\
            \textbf{Baseline} & A school is a place where one can learn. it is a place where one can get knowledge and skills \\ 
            \textbf{Decoder}  & a person who likes school also does. a person who likes school also does \\ 
            \textbf{Encoder} & enjoy learning is also a part of life of a person. a person who likes school also does enjoy learning \\
            \textbf{Attention.Q} & school is a place where people learn. a person who likes school also likes to learn. \\
        
            \textbf{LM Head + Att.Q} & school is a place where people learn. a person who likes school also does things like studying. \\
            \textbf{LayNorm + Att.Q} & school is a place where people learn. a person who likes school also does things like studying. \\
            \textbf{Self-Attention} &  a person who likes school also does things like studying. \\
            
        \end{tabular}
        }
        \end{tcolorbox}
        \begin{tcolorbox}[colback=blue!5!white, colframe=black,
        left=1mm,top=1mm,right=1mm,bottom=1mm, fonttitle=\small]
        \centering
        \resizebox{\columnwidth}{!}{
        \begin{tabular}{ p{8em} | p{23em} } 
            \textbf{Question}  & When someone is watching a film that is a documentary, what are they hoping to gain? \\ 
            \textbf{Answer Choices} & A) Like B) Insight C) Fear D) Being entertained E) Awareness \\
            \textbf{Answer} & Insight \\
            &  \\
            &  \textbf{Generated Explanation}  \\
            \textbf{Baseline} & insight is the ability to gain knowledge through observation. when someone watches a documentary, they are hoping to gain insight. \\ 
            \textbf{Decoder}  & insight is the ability to gain knowledge through observation. people often seek out information through watching a documentary. \\ 
            \textbf{Encoder} & insight is the ability to gain knowledge through observation. \\
            \textbf{Attention.Q} & insight is a kind of knowledge that is gained through observation. \\
            \textbf{LM Head + Att.Q} & insight is a quality that is gained when someone is watching a documentary. \\
            \textbf{LayNorm + Att.Q} & insight is gaining knowledge. people are hoping to gain knowledge by watching a documentary. \\
            \textbf{Self-Attention} &  insight is the ability to gain knowledge. when someone is watching a documentary, they are hoping to gain knowledge. \\
        \end{tabular}
        }
        \end{tcolorbox}
        \begin{tcolorbox}[colback=blue!5!white, colframe=black,
        left=1mm,top=1mm,right=1mm,bottom=1mm, fonttitle=\small]
        \centering
        \resizebox{\columnwidth}{!}{
        \begin{tabular}{ p{8em} | p{23em} } 
            \textbf{Question}  & When you go on vacation what do you have to do with things you want to take with you? \\ 
            \textbf{Answer Choices} & A) Pay bills. B) Save money. C) Pack. D) Sell. E) Have money \\
            \textbf{Answer} & Pack \\
            &  \\
            &  \textbf{Generated Explanation}  \\
            \textbf{Baseline} & packing means to take something with you. if you want to take something with you, you have to pack it. \\ 
            \textbf{Decoder}  & packing means packing to take with you when you go on vacation. if you want to take things with you, you have to pack them. \\ 
            \textbf{Encoder} & packing is the process of packing things you want to take with you. \\
            \textbf{Attention.Q} & packing is a process of taking things with you. when you go on vacation you have to pack things you want to take with you \\
            \textbf{LM Head + Att.Q} & when you go on vacation you have to pack things you want to take with you. \\
            \textbf{LayNorm + Att.Q} & tpacking is a process where you put things together to take with you when you go on vacation. \\
            \textbf{Self-Attention} &  you have to pack things when you go on vacation. packing means taking things with you. \\
        \end{tabular}
        }
        \end{tcolorbox}
    %\vspace{-1em}
    \caption{Examples of generated NLEs for ECQA.}
    \label{fig:examples_nle_ecqa}
\end{figure}

% Examples SBIC
\begin{figure}[H]
        \begin{tcolorbox}[colback=red!5!white, colframe=black,
        left=1mm,top=1mm,right=1mm,bottom=1mm, fonttitle=\small]
        \centering
        \resizebox{\columnwidth}{!}{
        \begin{tabular}{ p{8em} | p{23em} } 
            \textbf{Post}  & Watching conservacucks slobber all over the Western Wall makes me ill. \\ 
            \textbf{Label} & Offensive \\
            &  \\
            &  \textbf{Generated Explanation}  \\
            \textbf{Baseline} & this post implies that southerners are lazy. \\ 
            \textbf{Decoder}  & this post implies that conservacucks are slobbering \\ 
            \textbf{Encoder} & this post implies that conservacucks are disgusting \\
            \textbf{Attention.Q} & this post implies that conservacucks are slobbering all over the western wall. \\
            \textbf{LM Head + Att.Q} & tthis post implies that conservacucks are slobbering all over the western wall. \\
            \textbf{LayNorm + Att.Q} & this post makes fun of conservacucks\\
            \textbf{Self-Attention} &  this post implies that conservacucks are dirty. \\
        \end{tabular} 
        }
        \end{tcolorbox}
        \begin{tcolorbox}[colback=red!5!white, colframe=black,
        left=1mm,top=1mm,right=1mm,bottom=1mm, fonttitle=\small]
        \centering
        \resizebox{\columnwidth}{!}{
        \begin{tabular}{ p{8em} | p{23em} } 
            \textbf{Post}  & Whats the difference between a training gun and a police officer?  A training gun shoots blanks, an officer shoots blacks \\ 
            \textbf{Label} & Offensive \\
            &  \\
            &  \textbf{Generated Explanation}  \\
            \textbf{Baseline} & this post implies that black people are criminals \\ 
            \textbf{Decoder}  & this post implies that blacks are the only color that will be killed \\ 
            \textbf{Encoder} & this post implies that blacks are the victims of police brutality \\
            \textbf{Attention.Q} & this post implies that blacks are more dangerous \\
            \textbf{LM Head + Att.Q} & this post implies that blacks are more dangerous \\
            \textbf{LayNorm + Att.Q} & this post implies that black people are criminals \\
            \textbf{Self-Attention} &  this post implies that blacks are inferior \\
        \end{tabular} 
        }
        \end{tcolorbox}
        \begin{tcolorbox}[colback=red!5!white, colframe=black,
        left=1mm,top=1mm,right=1mm,bottom=1mm, fonttitle=\small]
        \centering
        \resizebox{\columnwidth}{!}{
        \begin{tabular}{ p{8em} | p{23em} } 
            \textbf{Post}  & What's the worst than a nerd being Basketball Player A Nigga Becoming a Doctor \\ 
            \textbf{Label} & Offensive \\
            &  \\
            &  \textbf{Generated Explanation}  \\
            \textbf{Baseline} & this post implies that black folks are less than white \\ 
            \textbf{Decoder}  & this post trivializes harm to victims \\ 
            \textbf{Encoder} & this post implies that blacks are worse than nerd \\
            \textbf{Attention.Q} & this post implies that blacks are better than nerds \\
            \textbf{LM Head + Att.Q} & this post implies that blacks are inferior \\
            \textbf{LayNorm + Att.Q} & this post implies that blacks are inferior \\
            \textbf{Self-Attention} &  this post implies that blacks are the only people who shoot training guns \\
            
        \end{tabular} 
        }
        \end{tcolorbox}
    %\vspace{-1em}
    \caption{Examples of generated NLEs for SBIC.}
    \label{fig:examples_nle_sbic}
\end{figure}

% Examples ComVE
\begin{figure}[H]
        \begin{tcolorbox}[colback=yellow!5!white, colframe=black,
        left=1mm,top=1mm,right=1mm,bottom=1mm, fonttitle=\small]
        \centering
        \resizebox{\columnwidth}{!}{
        \begin{tabular}{ p{8em} | p{23em} } 
            \textbf{Choice 1} & He used a towel to dry off after his shower. \\ 
            \textbf{Choice 2} & He used an apple to dry off after his shower. \\
            \textbf{Label} & Choice 2 \\ 
            &  \\
            &  \textbf{Generated Explanation}  \\
            \textbf{Baseline} & apple is a fruit, so it cannot dry off. \\ 
            \textbf{Decoder}  & apple is a fruit and does not dry off \\ 
            \textbf{Encoder} & an apple is not a very absorbent material \\
            \textbf{Attention.Q} & an apple doesn't dry off as well as a towel \\
            \textbf{LM Head + Att.Q} & an apple is a fruit and it doesn't dry off. \\
            \textbf{LayNorm + Att.Q} & an apple is a fruit and it's not a towel, so it's not drying off as well. \\
            \textbf{Self-Attention} &  apples are a fruit, and they don't dry well on their own. \\
        \end{tabular}
        }
        \end{tcolorbox}
        \begin{tcolorbox}[colback=yellow!5!white, colframe=black,
        left=1mm,top=1mm,right=1mm,bottom=1mm, fonttitle=\small]
        \centering
        \resizebox{\columnwidth}{!}{
        \begin{tabular}{ p{8em} | p{23em} } 
            \textbf{Choice 1} & He walked his fish \\ 
            \textbf{Choice 2} & He walked his dog \\
            \textbf{Label} & Choice 1 \\ 
            &  \\
            &  \textbf{Generated Explanation}  \\
            \textbf{Baseline} & fish have no legs \\ 
            \textbf{Decoder}  & fish cannot walk on water \\ 
            \textbf{Encoder} & fish cannot walk \\
            \textbf{Attention.Q} & fish are not able to walk \\
            \textbf{LM Head + Att.Q} & fish don't walk \\
            \textbf{LayNorm + Att.Q} & fish cannot walk \\
            \textbf{Self-Attention} &  fish cannot be walked \\
        \end{tabular}
        }
        \end{tcolorbox}
        \begin{tcolorbox}[colback=yellow!5!white, colframe=black,
        left=1mm,top=1mm,right=1mm,bottom=1mm, fonttitle=\small]
        \centering
        \resizebox{\columnwidth}{!}{
        \begin{tabular}{ p{8em} | p{23em} } 
            \textbf{Choice 1} & A ball is square and can roll \\ 
            \textbf{Choice 2} & A ball is round and can roll \\
            \textbf{Label} & Choice 1 \\ 
            &  \\
            &  \textbf{Generated Explanation}  \\
            \textbf{Baseline} & a square ball cannot roll \\ 
            \textbf{Decoder}  & square balls are not rollable \\ 
            \textbf{Encoder} & a ball is square cannot roll \\
            \textbf{Attention.Q} & balls are round and can roll \\
            \textbf{LM Head + Att.Q} & a ball is round and can roll. \\
            \textbf{LayNorm + Att.Q} & a square ball cannot roll \\
            \textbf{Self-Attention} &  a ball can roll only in a round shape \\
        \end{tabular}
        }
        \end{tcolorbox}
    %\vspace{-1em}
    \caption{Examples of generated NLEs for ComVE.}
    \label{fig:examples_nle_comve}
\end{figure}

%%%%%%%%%%%%%%%%%%%%%%%%%%%%%%%%%%%%%%%%%%%%%%%%%%%%%%%%
%%%%%%%%%%%%%%%%%%%%%%%%%%%%%%%%%%%%%%%%%%%%%%%%%%%%%%%%
%%%%%%%%%%%%%%%%%%%%%%%%%%%%%%%%%%%%%%%%%%%%%%%%%%%%%%%%

\section{\ours Full Results}\label{ap:sparse_full_results}

This section shows the results in terms of task accuracy, and NLEs quality all configurations of \ours and for different model sizes (\ie \texttt{T5-base}, \texttt{T5-large} and \texttt{T5-3b}). For each metric, we also break down the results by dataset. % to understand the unique traits of the generated NLEs. 

%%%%%%%%%%%%%%%%%%%%%%%%%%%%%%%%%%%%%%%%%%%%%%%%%%%%%%%%
%%%%%%%%%%%%%%%%%%%%%%%%%%%%%%%%%%%%%%%%%%%%%%%%%%%%%%%%
%%%%%%%%%%%%%%%%%%%%%%%%%%%%%%%%%%%%%%%%%%%%%%%%%%%%%%%%

%%%%%%%%%%%%%%%%%%%%%%%%%%%%%%%%
% Table comparison size
% Downstream, NLE quality and Compound evaluation summary (T5-large)
\begin{table*}[t]
    \centering
    \resizebox{0.9\textwidth}{!}{
    
    \begin{tabular}{c c c c c c c}
    \toprule
        %\textbf{Dataset} & \multicolumn{4}{c| }{Dataset} & \textbf{Avg} \\ \hline \hline
        % \midrule
        \textbf{\ours} & & ComVE & ECQA & SBIC & e-SNLI &  \textbf{Avg} \\ %\hline
        \midrule
        \multirow{2}{*}{\shortstack{LayerNorm +  Attention.Q \\ \texttt{T5-base}}} & Acc. &  53.22 {\tiny $\pm3.67$} $\triangledown$ &  39.35 {\tiny $\pm2.31$} $\triangledown$ &  62.11 {\tiny $\pm5.04$} $\triangledown$ &  72.63 {\tiny $\pm2.87$} $\triangledown$ &  56.83 {\tiny $\pm3.47$}  \\
           & nBERTs &  48.77 {\tiny $\pm3.37$} $\triangledown$ &     0.0 {\tiny $\pm0.0$} $\triangledown$ &  59.45 {\tiny $\pm5.47$} $\triangledown$ &  64.81 {\tiny $\pm3.13$} $\triangledown$ &  43.26 {\tiny $\pm2.99$}  \\
        \multirow{2}{*}{\shortstack{LayerNorm +  Attention.Q \\ \texttt{T5-large}}} & Acc. & \textgr{74.9} {\tiny $\pm5.3$} $\triangledown$ &  \textgr{55.8} {\tiny $\pm3.1$} $\triangledown$ &   \textgr{67.0} {\tiny $\pm4.4$} $\triangledown$&   82.6 {\tiny $\pm2.7$} $\triangledown$ &  \textgr{70.1} {\tiny $\pm3.9$} \\
            & nBERTs &   \textgr{69.0} {\tiny $\pm4.8$} &  \textgr{45.9} {\tiny $\pm3.7$} $\triangledown$ &   \textgr{64.3} {\tiny $\pm4.7$} &  75.6 {\tiny $\pm2.5$} $\triangledown$ &  \textgr{63.7} {\tiny $\pm3.9$} \\
        \multirow{2}{*}{\shortstack{LayerNorm +  Attention.Q \\ \texttt{T5-3B}}} & Acc. &   83.27 {\tiny $\pm4.52$} $\triangledown$ &  54.13 {\tiny $\pm3.86$} $\triangledown$ &  \textbf{68.87} {\tiny $\pm4.86$} $\triangledown$ &  79.16 {\tiny $\pm3.72$} $\triangledown$ &  71.36 {\tiny $\pm4.24$}  \\
           & nBERTs &   75.83 {\tiny $\pm4.14$} $\triangledown$ &  48.31 {\tiny $\pm3.46$} $\triangledown$ &  65.86 {\tiny $\pm5.07$} $\triangledown$ &  71.27 {\tiny $\pm3.44$} $\triangledown$ &  65.32 {\tiny $\pm4.03$}  \\
        \midrule

    \bottomrule
    \end{tabular}
    }
    \caption{Summary of best performing \ours configurations for \textit{LayerNorm +  Attention}. We report the average and the standard deviation over the 60 few-shot train-validation splits for the \textbf{accuracy} metric and the normalized BERTScore~(\textbf{nBERTs}).    
    % and \textbf{Compound Metric}~(Acc $\times$ BERTScore) metric. 
    In brackets are the percentages of fine-tuned weights for each \ours configuration. We show in \textbf{bold} the setting with the highest metric for each dataset, in \textbl{blue} the highest performance among \ours without considering the number of parameters, and in \textgr{green} the best-performing setting after considering the percentage of fine-tuned parameters. The trade-off between parameters and performances was computed using~$(1 - \% \text{params}) \times $~nBERTs). Significance testing was assessed via mean t-test compared with the baseline:~$\triangledown$~represents a p-value lower than $10^{-2}$.}
    \label{tab:sparse_lm+attq_sizes}
    \vspace{-3em}
\end{table*}

\subsection{Task Performance}\label{ap:sparse_task_full_results}

% t5-large all Datasets 
\begin{figure*}[]
    \centering
    \includegraphics[width=\textwidth]{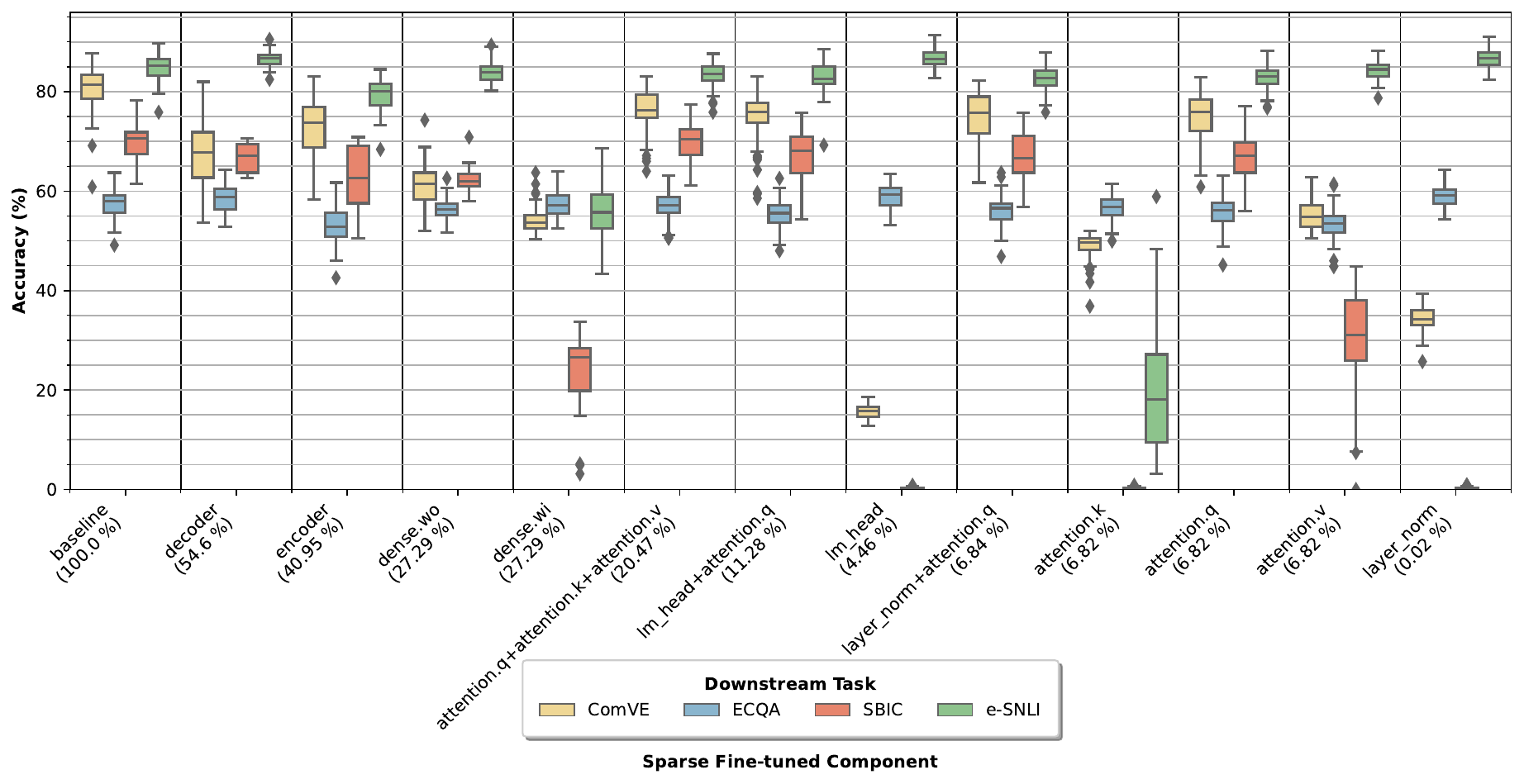}
    \caption{Distribution of the \textbf{accuracy} scores for different \ours configurations for \texttt{T5-large}.
    The percentage of parameters fine-tuned for each configuration is shown between brackets.}
    \label{fig:all_t5large_accuracy}
\end{figure*}

\begin{figure*}[]
    \centering
    \includegraphics[width=\textwidth]{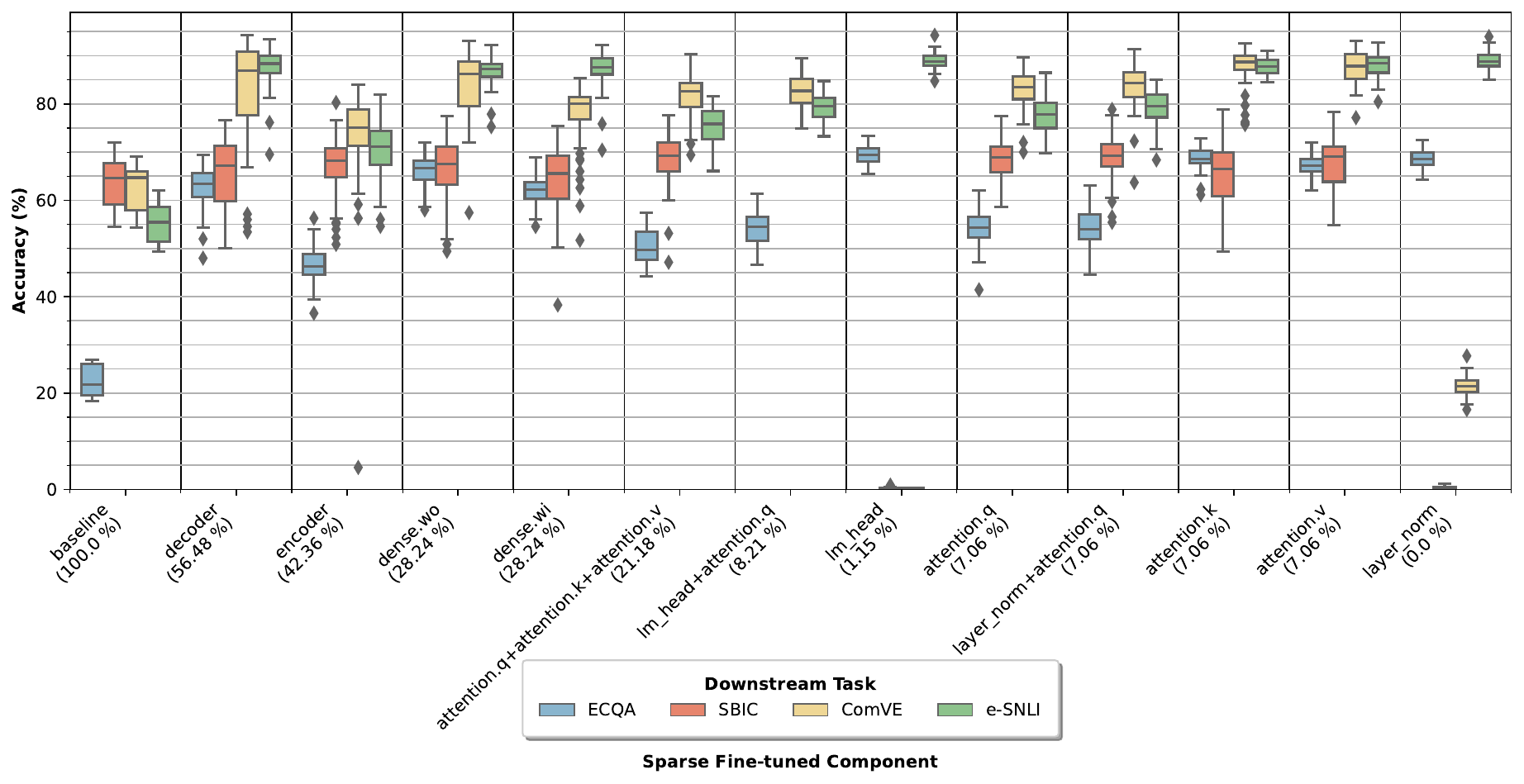}
    \caption{Distribution of the \textbf{accuracy} scores for different \ours configurations for \texttt{T5-3b}.
    The percentage of parameters fine-tuned for each configuration is shown between brackets.}
    \label{fig:all_t5-3b_accuracy}
\end{figure*}

\cref{fig:all_t5large_accuracy} depicts the distribution of the accuracy score for \ours configurations trained on top of \texttt{T5-large}. It can be observed that several \ours configurations exhibit similar performance as the baseline, particularly for ECQA and E-SNLI. The \ours configurations with the best task performance are \emph{Decoder}, \emph{Self-Attention KQV}, \emph{Self-attention Query}, and \emph{Layer Normalization}.  Remarkably, the \ours configurations do not show a higher variance than the baseline across the 60 train-validation splits (inter-quartile range). \cref{fig:all_t5-3b_accuracy} depicts the distribution of the accuracy score for \ours configurations trained on top of \texttt{T5-3b}. It can be observed that all \ours configurations outperform the baseline. However, the best performance for \texttt{T5-3b} is achieved by the sparse fine-tuning of the \emph{Self-attention Value Layer}. The results for \texttt{T5-base} can be observed in the breakdown done for each dataset. 

% t5-3b all Datasets 

\begin{figure*}[t]
    \centering
    \includegraphics[width=\textwidth]{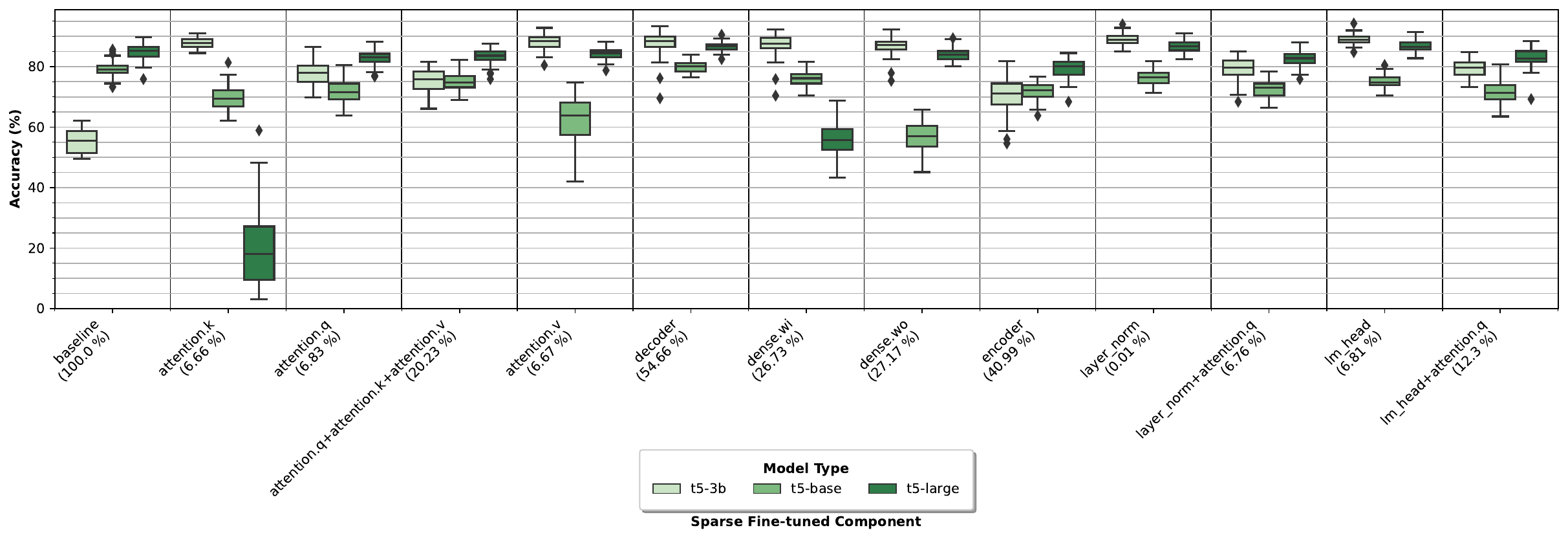}
    \caption{Distribution of the accuracies for different settings of \ours for the \textbf{e-SNLI} dataset. %The baseline model represents the work done by \citet{marasovic-etal-2022-shot}, where all the parameters of the LM were fine-tuned. 
    For each model, the variation represents the overall performance in each of the 60 train-validation splits. The percentage of parameters fine-tuned for each setup is depicted in brackets below the name of each configuration.}
    \label{fig:esnli_accuracy}
\end{figure*}

Figure~\ref{fig:esnli_accuracy} depicts the box plot with the distribution of the accuracy scores on e-SNLI for the 60 train-validation splits for different \ours configurations and the two pre-trained LM sizes. Overall, for e-SNLI, the task performance increases with the size of the model for most of the sparse fine-tuning configurations. Moreover, the interquartile range is considerably smaller when the model size increases (i.e., \texttt{T5-large} scores are less spread than the ones for \texttt{T5-base}). The highest median score was achieved by the fine-tuning of the \textit{Layer Normalization} in \texttt{T5-large}, followed very closely by the fine-tuning of the \textit{LM head} and the \textit{Decoder} in \texttt{T5-large}. The combination of components (\ie Layer Norm + Self-attention Query) performed very closely to the best-performing settings. 

% e-SNLI Plots 

\begin{figure*}[t]
    \centering
    \includegraphics[width=\textwidth]{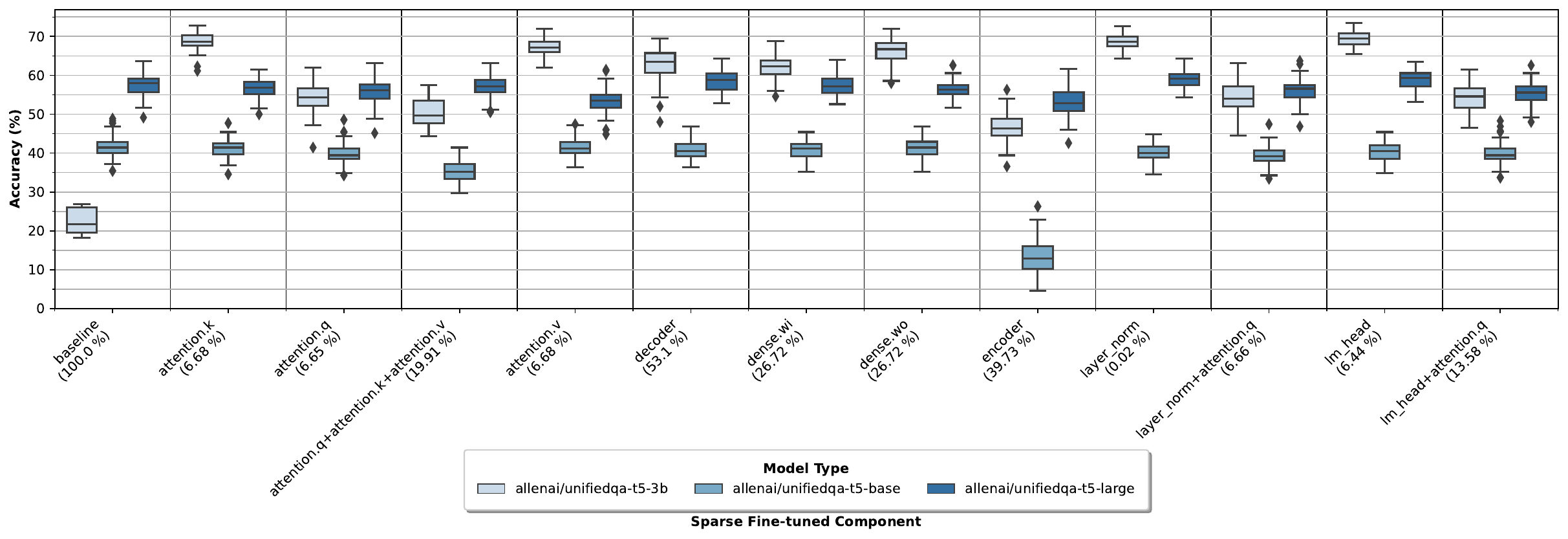}
    \caption{Distribution of the accuracy scores for different \ours settings for the \textbf{ECQA} dataset. %The baseline model represents the work done by \citet{marasovic-etal-2022-shot}, where all the parameters of the LM were fine-tuned. 
    For each model, the variation represents the overall performance in each of the 60 train-validation splits. The percentage of parameters fine-tuned for each setup is depicted in brackets below the name of each configuration.todo{Update plot with t5-3b results}}
    \label{fig:ecqa_accuracy}
\end{figure*}

For the ECQA dataset, Figure~\ref{fig:ecqa_accuracy} shows the box plot with the accuracy scores for different \ours setups. It can be observed that the performance of the larger LM (i.e., \texttt{T5-large}) is consistently better than \texttt{T5-base}. Overall, the accuracy is fairly similar for all the \ours configurations for a given LM size, with an average of $58\%$ and $42\%$ for \texttt{T5-large} and \texttt{T5-base}, respectively. Note that the random guess accuracy is $20\%$ for the ECQA dataset, since there are 5 possible answer choices. The highest accuracy was achieved by the fine-tuning of the \textit{Decoder} in \texttt{T5-large}, followed very closely by the fine-tuning of the \textit{Layer Normalization} and \emph{LM Head}. The combination of components achieves a slightly lower performance than single components for the task prediction. Surprisingly, for ECQA, the variability for a given combination of configuration-model (\ie each box) is higher for \texttt{T5-large} than for \texttt{T5-base}. Moreover, the fine-tuning of the \textit{Encoder} for \texttt{T5-base} gives worse results in comparison with all the other configurations. Besides the setting where only the \textit{Encoder} is fine-tuned for \texttt{T5-base}, the highest observed range in ECQA is roughly $14\%$.

% ECQA Plots 

\begin{figure*}[t]
    \centering
    \includegraphics[width=\textwidth]{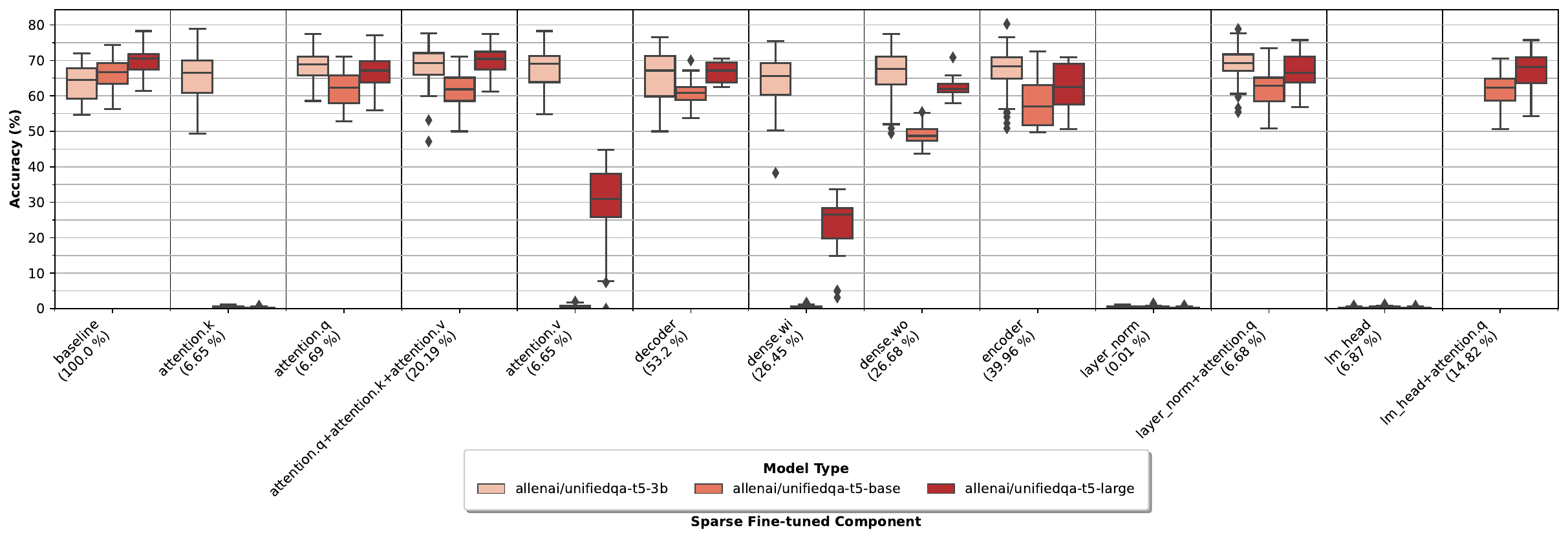}
    \caption{Distribution of the accuracy scores for different settings of \ours for the \textbf{SBIC} dataset. %The baseline model represents the work done by \citet{marasovic-etal-2022-shot}, where all the parameters of the LM were fine-tuned. 
    For each model, the variation represents the overall performance in each of the 60 train-validation splits. The percentage of parameters fine-tuned for each setup is depicted in brackets below the name of each configuration.}
    \label{fig:sbic_accuracy}
\end{figure*}

For the SBIC dataset, \cref{fig:sbic_accuracy} depicts the box-plot with the dispersion of accuracy scores for \texttt{T5-base} and \texttt{T5-large}. Recall that for the SBIC dataset, we fine-tune the \texttt{UnifiedQA} variant of T5. In general, it can be seen that the accuracy score surges when the model size is increased; thus, the best accuracy scores for a given sparse fine-tuning setup are found for the \texttt{T5-large}. The best median accuracy performance is achieved by the baseline. However, the difference in the median scores between the best and the second and third best-ranked configurations (\ie \textit{Self-attention Layer} and \textit{Layer Normalization + Self-attention Query}, respectively) are less than $3\%$. 
The maximum variance among scores for the 3 best-performing \ours configurations is roughly $15\%$. Furthermore, it can be observed that for many very sparse fine-tuning configurations, the accuracy score is close to or equal to zero. Even though the performance of a random model is $50\%$, an accuracy of $0\%$ is feasible in our scenario as the model could generate different words from the ones expected as labels. In this regard, the accuracy scores of zero are a consequence of the fact that, after the conditional generation, the model generates neither \textit{``offensive"} nor \textit{``non-offensive"} for any sample in the validation set. Notice that this phenomenon is particularly happening when only a small fraction of weights is fine-tuned. 

% SBIC Plots 

% ComVE Plots 
\begin{figure*}[t]
    \centering
    \includegraphics[width=\textwidth]{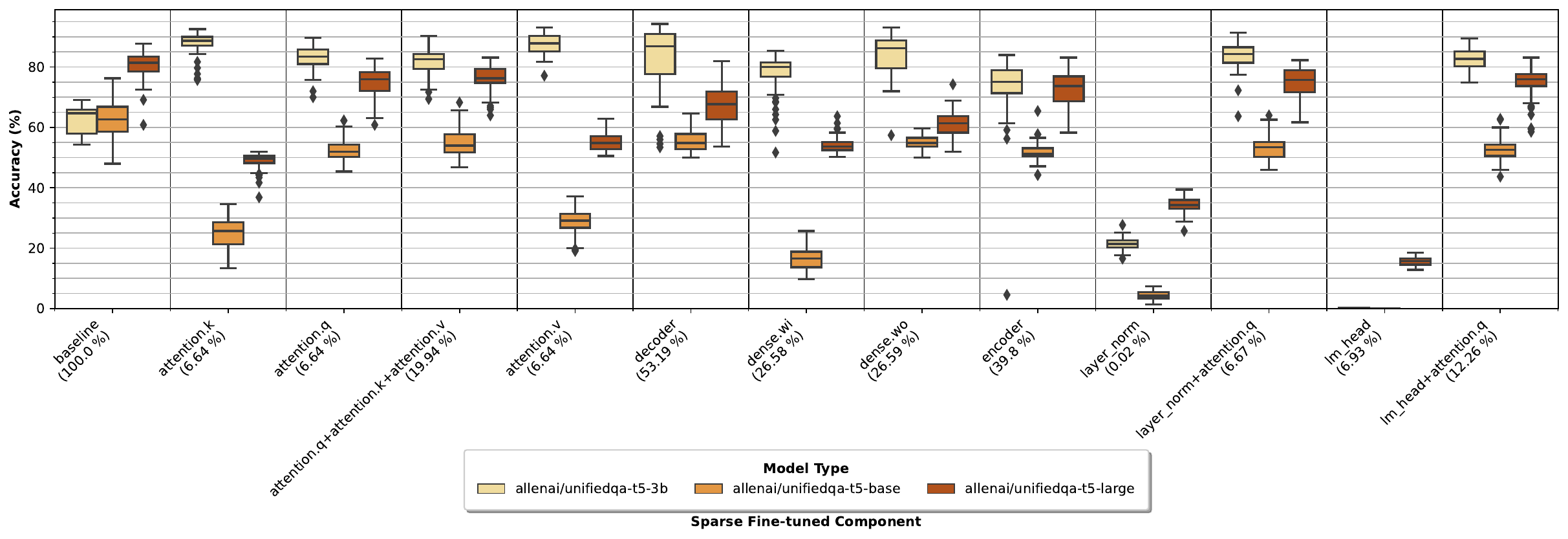}
    \caption{Distribution of the accuracy scores for different settings of \ours for the \textbf{ComVE} dataset. %The baseline model represents the work done by \citet{marasovic-etal-2022-shot}, where all the parameters of the LM were fine-tuned. 
    For each model, the variation represents the overall performance in each of the 60 train-validation splits. The percentage of parameters fine-tuned for each setup is depicted in brackets below the name of each configuration.}
    \label{fig:comve_accuracy}
\end{figure*}

For the ComVE dataset, we show in \cref{fig:comve_accuracy} the accuracy for the 60 different train-validation splits for different \ours settings and model sizes. It can be seen that the best-performing setting in terms of accuracy is the baseline for \texttt{UNIFIEDQA-T5-large}. (\ie \textit{Self-attention Layer} and \textit{Layer Normalization + Self-attention Query} fine-tuning are the second and third best performing, respectively. Overall, the fine-tuning of the \textit{Normalization Layer} leads to the worst task performance. Moreover, it can be observed that the performance increases with the size of the model, thus \texttt{UNIFIEDQA-T5-large} always performs better than \texttt{UNIFIEDQA-T5-base} for all the fine-tuning configurations. The smallest gap in performance between model sizes (\texttt{UNIFIEDQA-T5-large} vs.\ \texttt{UNIFIEDQA-T5-base}) happens for the fine-tuning of the \textit{Dense Layer}. Conversely, the maximum spread in performance (\ie the difference between the best and the worst split) is around $21\%$ for models trained using the \texttt{UNIFIEDQA-T5-large} architecture.

%%%%%%%%%%%%%%%%%%%%%%%%%%%%%%%%%%%%%%%%%%%%%%%%%%%%%%%%
%%%%%%%%%%%%%%%%%%%%%%%%%%%%%%%%%%%%%%%%%%%%%%%%%%%%%%%%
%%%%%%%%%%%%%%%%%%%%%%%%%%%%%%%%%%%%%%%%%%%%%%%%%%%%%%%%

\subsection{Explanation Generation Performance}\label{ap:sparse_explanations_full_results}

\cref{fig:all_t5large_explanations} shows the box-plot with the normalized BERTscores for different \ours setups fine-tuned on top of \texttt{T5-large}. In addition to explained in the main text, it can be seen that combinations of components lead to less variance in the score achieved for the 60 train-test splits (see the interquartile range). Furthermore,~\cref{tab:sparse_summary_all_t5-3b} shows the performance summary for the downstream performance and the NLEs quality for \texttt{T5-3b}. It can be observed that the \emph{Attention Value Layer} achieves the best performance on average. We highlight that \ours outperforms the baseline (\ie full fine-tuning) for all datasets.

% % Explanations t5-large all Datasets 
% \begin{figure*}[t]
%     \centering
%     \includegraphics[width=\textwidth]{figures/sparse_finetuning/all-t5-large-dev_bertscore_correct_normalized-horizontal-no_blocks.pdf}
%     \caption{Distribution of the \textbf{normalized BERTScore} for different \ours settings of sparse fine-tuning for \texttt{T5-large}. %The baseline is the model proposed by \citet{marasovic-etal-2022-shot}, where all the parameters were fine-tuned. 
%     The percentage of fine-tuned parameters is shown between brackets.}
%     \label{fig:all_t5large_explanations}
% \end{figure*}

% Downstream, NLE quality and Compound evaluation summary (T5-large)
\begin{table*}[t]
    \centering
    \resizebox{0.9\textwidth}{!}{
    
    \begin{tabular}{c c c c c c c}
    \toprule
        %\textbf{Dataset} & \multicolumn{4}{c| }{Dataset} & \textbf{Avg} \\ \hline \hline
        % \midrule
        \textbf{\ours} & & ComVE & ECQA & SBIC & e-SNLI &  \textbf{Avg} \\ %\hline
        \midrule
        \multirow{2}{*}{\shortstack{Baseline \\ ($100.00\%$)}} & Acc. &                  62.48 {\tiny $\pm6.03$}  &                 22.39 {\tiny $\pm3.61$}  &                 63.55 {\tiny $\pm6.59$}  &                  55.3 {\tiny $\pm4.98$}  &   50.93 {\tiny $\pm5.3$}  \\
                   & nBERTs &                   55.55 {\tiny $\pm5.6$}  &                 19.73 {\tiny $\pm3.22$}  &                 61.21 {\tiny $\pm6.79$}  &                 49.25 {\tiny $\pm4.36$}  &  46.44 {\tiny $\pm4.99$}  \\
        \midrule
        \multirow{2}{*}{\shortstack{Decoder \\ ($54.60\%$)}} & Acc. &  83.67 {\tiny $\pm10.12$} $\triangledown$ &  62.62 {\tiny $\pm4.16$} $\triangledown$ &                 65.59 {\tiny $\pm7.51$}  &  87.48 {\tiny $\pm4.02$} $\triangledown$ &  74.84 {\tiny $\pm6.45$}  \\
                   & nBERTs &   74.66 {\tiny $\pm9.02$} $\triangledown$ &  55.31 {\tiny $\pm3.65$} $\triangledown$ &                 62.72 {\tiny $\pm7.66$}  &   77.92 {\tiny $\pm3.7$} $\triangledown$ &  67.65 {\tiny $\pm6.01$}  \\
        \midrule
        \multirow{2}{*}{\shortstack{Encoder \\ ($40.95\%$)}} & Acc. &  73.14 {\tiny $\pm11.24$} $\triangledown$ &  46.23 {\tiny $\pm3.96$} $\triangledown$ &                 66.81 {\tiny $\pm6.43$}  &  70.34 {\tiny $\pm5.79$} $\triangledown$ &  64.13 {\tiny $\pm6.86$}  \\
                   & nBERTs &   66.7 {\tiny $\pm10.28$} $\triangledown$ &  41.46 {\tiny $\pm3.56$} $\triangledown$ &                  64.45 {\tiny $\pm6.7$}  &  63.79 {\tiny $\pm5.28$} $\triangledown$ &   59.1 {\tiny $\pm6.46$}  \\
        \midrule
        \multirow{2}{*}{\shortstack{Dense.wo \\ ($27.29\%$)}} & Acc. &   83.91 {\tiny $\pm6.54$} $\triangledown$ &  66.21 {\tiny $\pm3.12$} $\triangledown$ &                 66.64 {\tiny $\pm6.46$}  &   86.85 {\tiny $\pm3.0$} $\triangledown$ &   75.9 {\tiny $\pm4.78$}  \\
                   & nBERTs &    76.1 {\tiny $\pm6.04$} $\triangledown$ &  59.12 {\tiny $\pm2.76$} $\triangledown$ &                 63.87 {\tiny $\pm6.51$}  &  78.24 {\tiny $\pm2.78$} $\triangledown$ &  69.33 {\tiny $\pm4.52$}  \\
        \midrule
        \multirow{2}{*}{\shortstack{Dense.wi \\ ($27.29\%$)}} & Acc. &    77.6 {\tiny $\pm6.63$} $\triangledown$ &  62.12 {\tiny $\pm2.75$} $\triangledown$ &                  63.99 {\tiny $\pm7.4$}  &   87.31 {\tiny $\pm3.6$} $\triangledown$ &   72.76 {\tiny $\pm5.1$}  \\
                   & nBERTs &   70.21 {\tiny $\pm6.04$} $\triangledown$ &  55.12 {\tiny $\pm2.44$} $\triangledown$ &                 61.05 {\tiny $\pm7.43$}  &  78.24 {\tiny $\pm3.28$} $\triangledown$ &   66.16 {\tiny $\pm4.8$}  \\
        \midrule
        \multirow{2}{*}{\shortstack{Attention KQV \\ ($20.47\%$)}} & Acc. &   81.73 {\tiny $\pm4.14$} $\triangledown$ &  50.24 {\tiny $\pm3.48$} $\triangledown$ &  68.84 {\tiny $\pm5.37$} $\triangledown$ &   75.3 {\tiny $\pm3.78$} $\triangledown$ &  69.03 {\tiny $\pm4.19$}  \\
                   & nBERTs &   74.27 {\tiny $\pm3.84$} $\triangledown$ &  44.79 {\tiny $\pm3.06$} $\triangledown$ &  \textbf{66.12} {\tiny $\pm5.46$} $\triangledown$ &  67.67 {\tiny $\pm3.36$} $\triangledown$ &  63.21 {\tiny $\pm3.93$}  \\
        \midrule
        \multirow{2}{*}{\shortstack{LM head +  Attention.Q \\ ($11.28\%$)}} & Acc. &   82.59 {\tiny $\pm3.37$} $\triangledown$ &  54.28 {\tiny $\pm3.57$} $\triangledown$ &                    0.0 {\tiny $\pm0.0$}  &  79.33 {\tiny $\pm2.94$} $\triangledown$ &  72.07 {\tiny $\pm3.29$}  \\
                   & nBERTs &     75.2 {\tiny $\pm3.0$} $\triangledown$ &  48.42 {\tiny $\pm3.17$} $\triangledown$ &                    0.0 {\tiny $\pm0.0$}  &  71.52 {\tiny $\pm2.74$} $\triangledown$ &  65.05 {\tiny $\pm2.97$}  \\
        \midrule
        \multirow{2}{*}{\shortstack{LM head \\ ($4.46\%$)}} & Acc. &    0.09 {\tiny $\pm0.13$} $\triangledown$ &  \textbf{69.43} {\tiny $\pm1.88$} $\triangledown$ &   0.23 {\tiny $\pm0.22$} $\triangledown$ &  \textbf{89.04} {\tiny $\pm1.63$} $\triangledown$ &   39.7 {\tiny $\pm0.96$}  \\
                   & nBERTs &      0.0 {\tiny $\pm0.0$} $\triangledown$ &     0.0 {\tiny $\pm0.0$} $\triangledown$ &   0.19 {\tiny $\pm0.18$} $\triangledown$ &     0.0 {\tiny $\pm0.0$} $\triangledown$ &   0.05 {\tiny $\pm0.04$}  \\
        \midrule
        \multirow{2}{*}{\shortstack{LayerNorm +  Attention.Q \\ ($6.84\%$)}} & Acc. &   83.27 {\tiny $\pm4.52$} $\triangledown$ &  54.13 {\tiny $\pm3.86$} $\triangledown$ &  \textbf{68.87} {\tiny $\pm4.86$} $\triangledown$ &  79.16 {\tiny $\pm3.72$} $\triangledown$ &  71.36 {\tiny $\pm4.24$}  \\
                   & nBERTs &   75.83 {\tiny $\pm4.14$} $\triangledown$ &  48.31 {\tiny $\pm3.46$} $\triangledown$ &  65.86 {\tiny $\pm5.07$} $\triangledown$ &  71.27 {\tiny $\pm3.44$} $\triangledown$ &  65.32 {\tiny $\pm4.03$}  \\
        \midrule
        \multirow{2}{*}{\shortstack{Attention.Q \\ ($6.82\%$)}} & Acc. &   83.09 {\tiny $\pm4.15$} $\triangledown$ &  54.39 {\tiny $\pm3.66$} $\triangledown$ &  68.44 {\tiny $\pm4.44$} $\triangledown$ &  77.88 {\tiny $\pm3.66$} $\triangledown$ &  70.95 {\tiny $\pm3.98$}  \\
                   & nBERTs &   75.65 {\tiny $\pm3.76$} $\triangledown$ &  48.56 {\tiny $\pm3.24$} $\triangledown$ &   65.4 {\tiny $\pm4.68$} $\triangledown$ &  70.23 {\tiny $\pm3.41$} $\triangledown$ &  64.96 {\tiny $\pm3.77$}  \\
        \midrule
        \multirow{2}{*}{\shortstack{Attention.K \\ ($6.82\%$)}} & Acc. &    87.7 {\tiny $\pm3.83$} $\triangledown$ &  68.74 {\tiny $\pm2.29$} $\triangledown$ &                 65.48 {\tiny $\pm6.26$}  &   87.8 {\tiny $\pm1.83$} $\triangledown$ &  77.43 {\tiny $\pm3.55$}  \\
                   & nBERTs &   \textbf{80.01} {\tiny $\pm3.52$} $\triangledown$ &  \textbf{61.25} {\tiny $\pm2.07$} $\triangledown$ &                  62.41 {\tiny $\pm6.5$}  &  79.55 {\tiny $\pm1.62$} $\triangledown$ &   70.8 {\tiny $\pm3.43$}  \\
        \midrule
        \multirow{2}{*}{\shortstack{Attention.V \\ ($6.82\%$)}} & Acc. &   \textbf{87.72} {\tiny $\pm3.16$} $\triangledown$ &  67.22 {\tiny $\pm2.14$} $\triangledown$ &  68.11 {\tiny $\pm5.19$} $\triangledown$ &  88.17 {\tiny $\pm2.38$} $\triangledown$ &  \textbf{77.81} {\tiny $\pm3.22$}  \\
                   & nBERTs &   79.87 {\tiny $\pm2.92$} $\triangledown$ &   60.12 {\tiny $\pm1.9$} $\triangledown$ &  65.67 {\tiny $\pm5.09$} $\triangledown$ &  \textbf{79.63} {\tiny $\pm2.26$} $\triangledown$ &  \textbf{71.32} {\tiny $\pm3.04$}  \\
        \midrule
        \multirow{2}{*}{\shortstack{LayerNorm \\ ($0.02\%$)}} & Acc. &   21.37 {\tiny $\pm2.06$} $\triangledown$ &  68.71 {\tiny $\pm1.89$} $\triangledown$ &   0.29 {\tiny $\pm0.27$} $\triangledown$ &  88.91 {\tiny $\pm1.74$} $\triangledown$ &  44.82 {\tiny $\pm1.49$}  \\
                   & nBERTs &      0.0 {\tiny $\pm0.0$} $\triangledown$ &     0.0 {\tiny $\pm0.0$} $\triangledown$ &   0.24 {\tiny $\pm0.22$} $\triangledown$ &     0.0 {\tiny $\pm0.0$} $\triangledown$ &   0.06 {\tiny $\pm0.06$}  \\
        \bottomrule
    \end{tabular}
    
    }
    \caption{Summary of best performing \ours configurations for \texttt{T5-3B}. We report the average and the standard deviation over the 60 few-shot train-validation splits for the \textbf{accuracy} metric and the normalized BERTScore~(\textbf{nBERTs}). In brackets are the percentages of fine-tuned weights for each \ours configuration. We show in \textbf{bold} the setting with the highest metric for each dataset. Significance testing was assessed via mean t-test in comparison with the baseline: $\triangledown$~represents a p-value lower than $10^{-2}$.}
    \label{tab:sparse_summary_all_t5-3b}
\end{table*}

\begin{figure*}[!h]
    \centering
    \includegraphics[width=\textwidth]{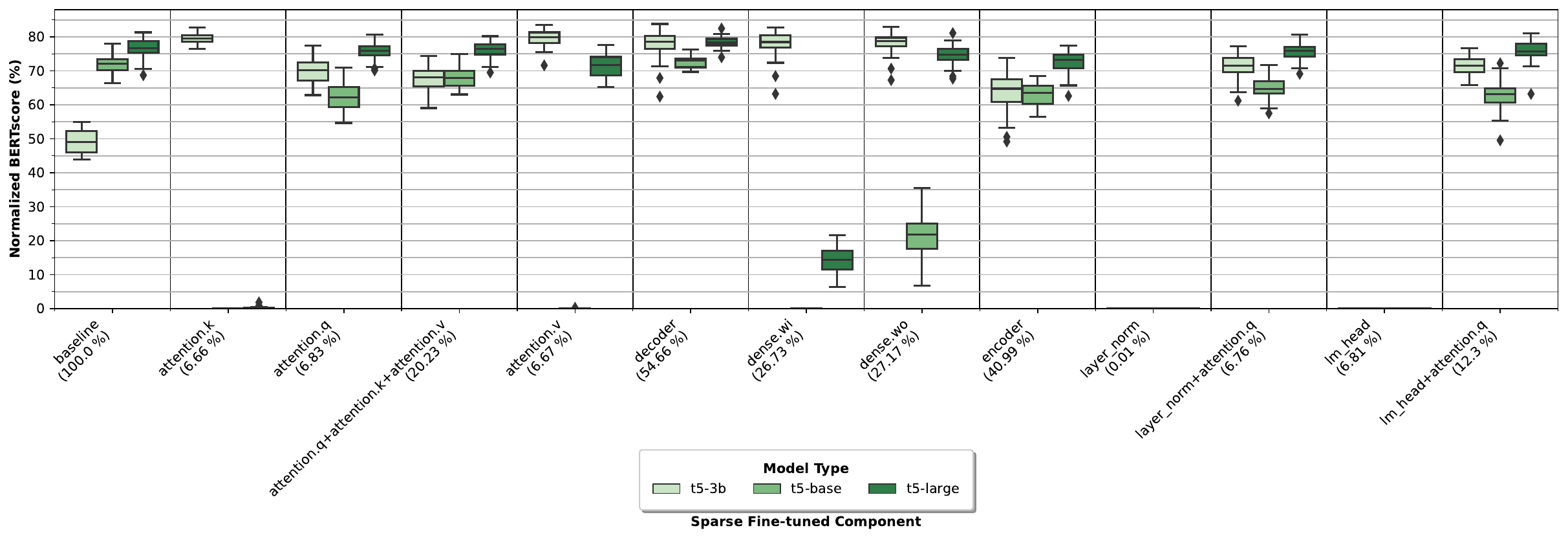}
    \caption{Distribution of the \textbf{normalized BERTscore} for different settings of sparse fine-tuning for the \textbf{e-SNLI} dataset. %The baseline model represents the work done by~\citet{marasovic-etal-2022-shot}, where all the parameters of the LM were fine-tuned. 
    For each model, the box represents the overall performance over the 60 train-validation splits. The percentage of parameters fine-tuned for each setup is depicted in brackets below the name of each configuration.}
    \label{fig:esnli_bertscore}
\end{figure*}

For e-SNLI, \cref{fig:esnli_bertscore} shows the normalized BERTscore over the 60 few-shot learning splits for different \ours configurations. Overall, for every sparse fine-tuning setting, the BERTscore is consistently higher for the largest PLM (\ie \texttt{T5-large}). However, the gap in performance is smaller for the best-performing sparse fine-tuning configurations. For instance, the difference in the average normalized BERTscore values between \texttt{T5-large} and \texttt{T5-base} for the best performing \ours~(i.e., \textit{Decoder}) is roughly $5\%$ while for the worst performing configuration is around $68\%$. The first five best-performing \ours configurations for \texttt{T5-large} are \textit{Decoder}, \textit{Baseline}, \textit{Self-attention KVQ}, \textit{Layer Normalization + Self-attention Q}, and \textit{Self-attention Values}. Note that the normalized BERTscore is zero for some sparse fine-tuning configurations (e.g., \textit{Layer Normalization}). This is mostly happening when the sparse fine-tuning is applied to small models (i.e., \texttt{T5-base}). The fact that the BERTscore is zero for a given configuration for all the samples in a split implies that the generated NLEs are always empty. We explore the reasons behind this phenomenon in Section~\ref{sec:discussion}

% Explanations t5-3b all Datasets 
\begin{figure*}[t]
    \centering
    \includegraphics[width=\textwidth]{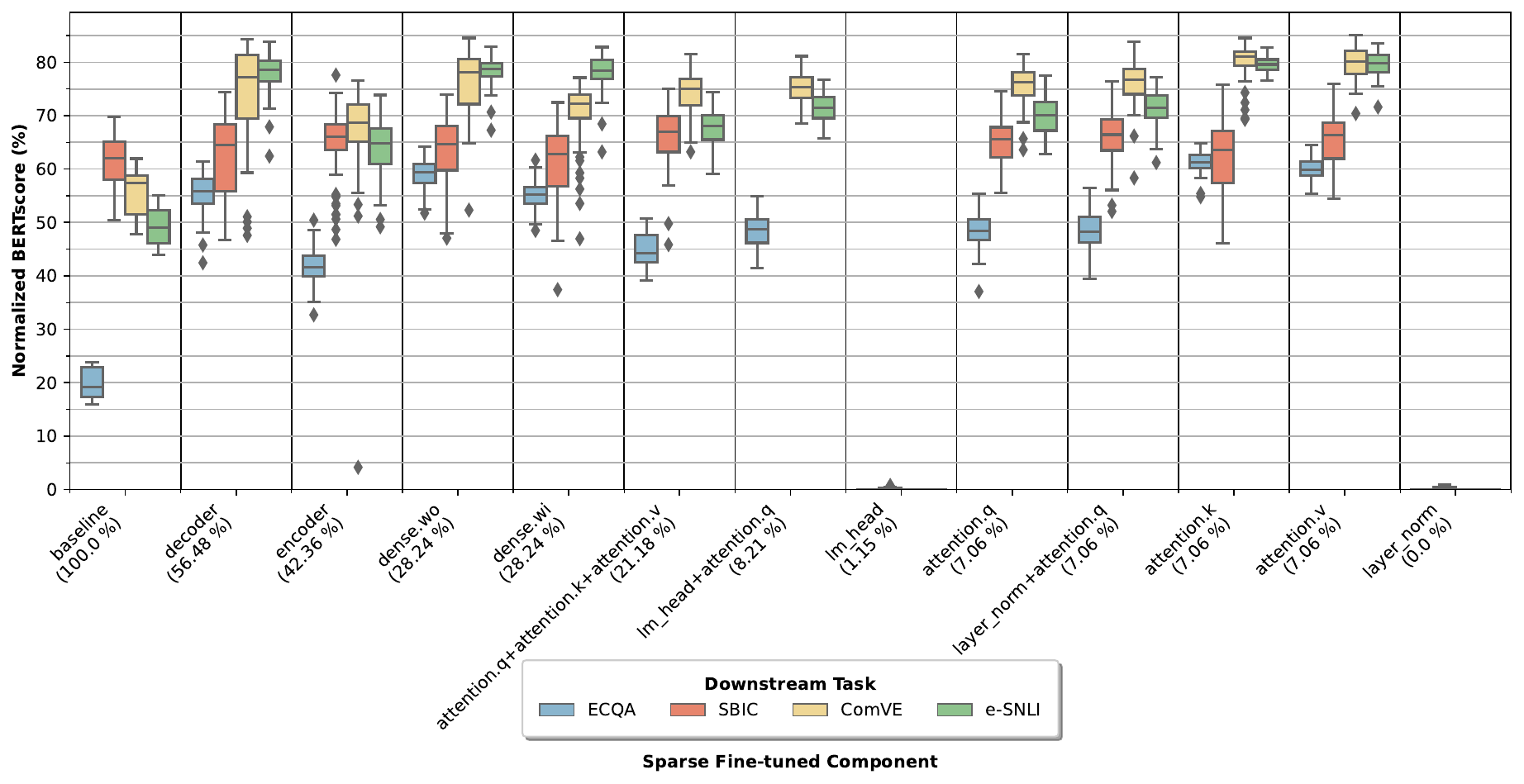}
    \caption{Distribution of the \textbf{normalized BERTScore} for different \ours settings of sparse fine-tuning for \texttt{T5-3b}. The percentage of fine-tuned parameters is shown between brackets.}
    \label{fig:all_t53b_explanations}
\end{figure*}

% e-SNLI Plots 

\begin{figure*}[!h]
    \centering
    \includegraphics[width=\textwidth]{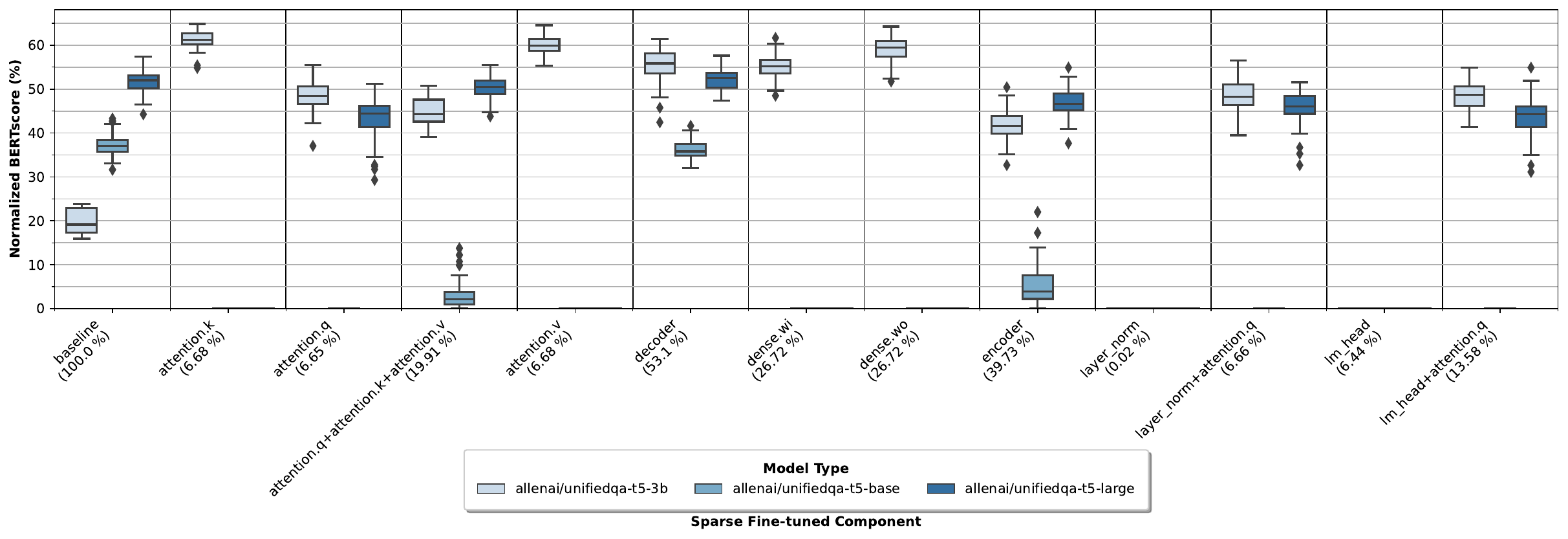}
    \caption{Distribution of the \textbf{normalized BERTscore} for different settings of sparse fine-tuning for the \textbf{ECQA} dataset. %The baseline model represents the work done by~\citet{marasovic-etal-2022-shot}, where all the parameters of the LM were fine-tuned. 
    For each model, the box represents the overall performance over the 60 train-validation splits. The percentage of parameters fine-tuned for each setup is depicted in brackets below the name of each configuration.}
    \label{fig:ecqa_bertscore}
\end{figure*}

For the ECQA dataset, we show in \cref{fig:ecqa_bertscore} the spread of the normalized BERTscore for all \ours configurations. Without exception, the largest model (\texttt{T5-large}) outperforms the \texttt{T5-base} models for every setting. Remarkably, for ECQA, many sparse fine-tuning configurations lead to the generation of empty explanations. Particularly, only the fine-tuning of the \textit{Baseline}, the \textit{Decoder}, and the \textit{Encoder} are able to consistently generate non-empty explanations no matter the size of the model. Among the configurations that generate non-empty explanations, the best normalized BERTscores are achieved by the \textit{Decoder} sparse fine-tuning, followed by the \textit{Baseline} and \textit{Encoder Blocks} fine-tuning. Note that for all of these configurations, the interquartile range is smaller than $6\%$ regardless of the model size. Moreover, the fine-tuning of \textit{Self-attention Query} achieves competitive results for \texttt{T5-large} but zero BERTscore for \texttt{T5-base}.

% ECQA Plots 

\begin{figure*}[!h]
    \centering
    \includegraphics[width=\textwidth]{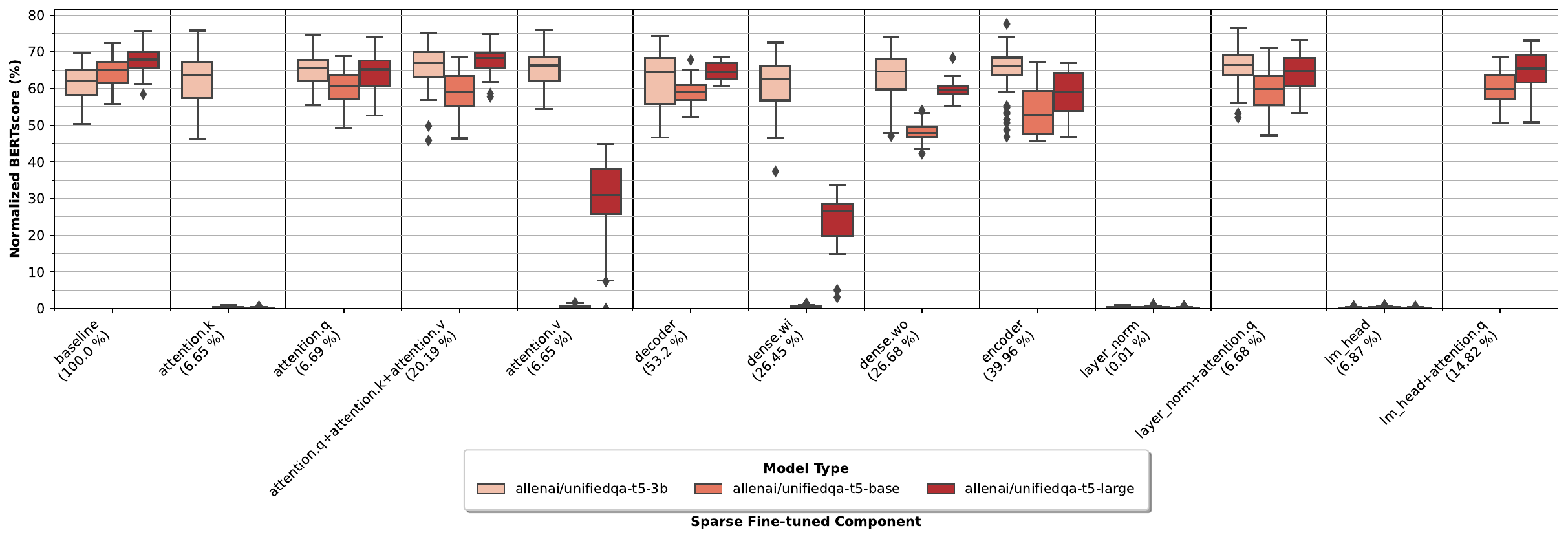}
    \caption{Distribution of the \textbf{normalized BERTscore} for different settings of sparse fine-tuning for the \textbf{SBIC} dataset. %The baseline model represents the work done by~\citet{marasovic-etal-2022-shot}, where all the parameters of the LM were fine-tuned. 
    For each model, the box represents the overall performance over the 60 train-validation splits. The percentage of parameters fine-tuned for each setup is depicted in brackets below the name of each configuration.}
    \label{fig:sbic_bertscore}
\end{figure*}

\cref{fig:sbic_bertscore} shows the normalized BERTscore results for the SBIC dataset. Recall that for the SBIC dataset, we fine-tune the \texttt{UnifiedQA} variant of T5. As expected, the model size contributes to better performance. Consequently, the BERTscore is higher for the \texttt{T5-large} model for every sparse fine-tuning configuration. The best BERTscore median is achieved by the \textit{Baseline} in combination with the \texttt{UNIFIEDQA-T5-large}, with a metric value of $\approx68\%$. The second and third best-performing setups are the \textit{Decoder} and the \textit{Encoder}, respectively. Moreover, the fine-tuning of layers such as the \textit{Normalization Layer} or \textit{Self-attention Layer} results in the generation of text that does not contain the explanation token ``because'', thus the BERTscore is close to zero for those configurations.

% SBIC Plots 

\begin{figure*}[!h]
    \centering
    \includegraphics[width=\textwidth]{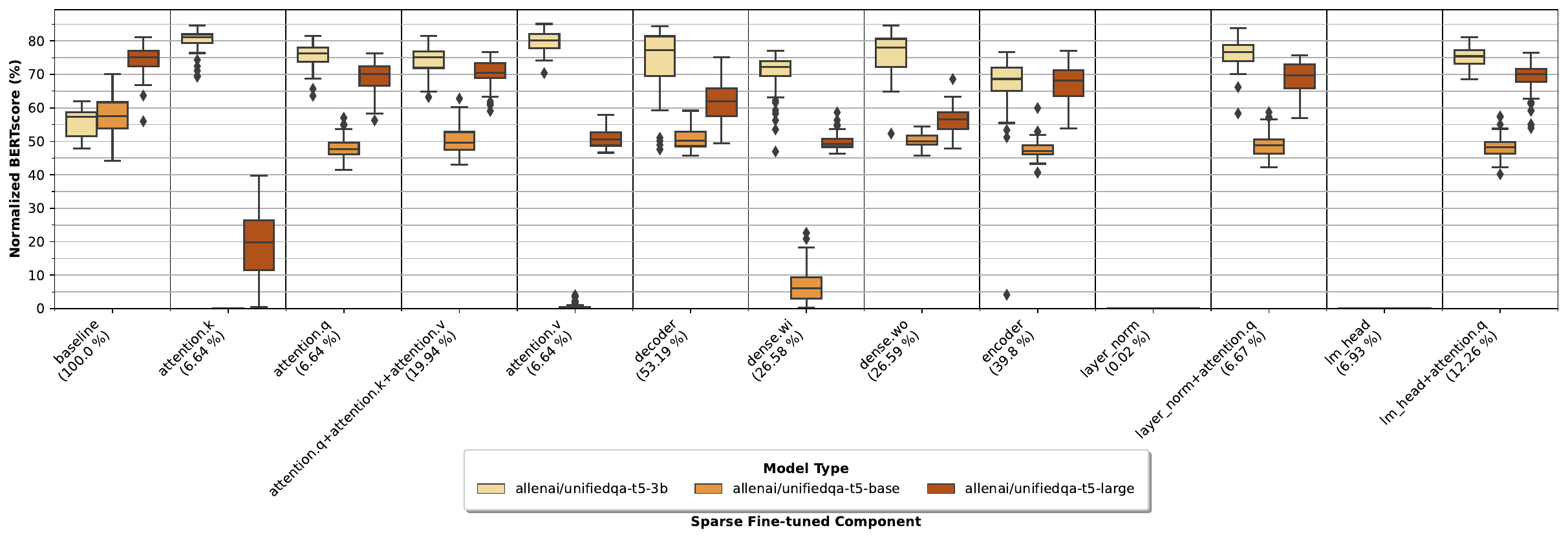}
    \caption{Distribution of the \textbf{normalized BERTscore} for different settings of sparse fine-tuning for the \textbf{ComVE} dataset. The baseline model represents the work done by \cite{marasovic-etal-2022-shot}, where all the parameters of the LM were fine-tuned. For each model, the box represents the overall performance over the 60 train-validation splits. The percentage of parameters fine-tuned for each setup is depicted in brackets below the name of each configuration.}
    \label{fig:comve_bertscore}
\end{figure*}

We depict in \cref{fig:comve_bertscore} the variation of the normalized BERTscore metric over the 60 different train-validation splits for the \ours configurations. Recall that for ComVE dataset, we fine-tune the \texttt{UnifiedQA} variant of T5. Overall, the BERTscore is substantially higher for \texttt{T5-large}. The best BERTscore for \texttt{T5-large} is obtained by the \textit{Baseline} fine-tuning, with a median score of $75\%$ for the 60 different seeds. Similar behavior can be seen for \texttt{T5-base}, where \textit{Baseline} is also the setting with the best explanations (from the perspective of the automatic metric). The second and third best sparse fine-tuning setups are the \textit{Self-attention Query} and \textit{Baseline}, respectively. Notice that the difference in the median between the \textit{Baseline} and the \textit{Encoder} is around $3\%$. Moreover, the variance among the different splits for a given sparse fine-tuning setting is on average higher than for the \textit{Baseline}. Remarkably, the sparse fine-tuning over the \textit{Normalization Layer} was the only setting that obtained a zero BERTscore for the ComVE dataset.

% ComVE Plots 

\subsection{Other PEFT Baselines}\label{ap:further_peft_exploration}

In order to make our approach comparable in the number of parameters, we test LoRa~\cite{hu2022lora} using higher ranks. \cref{tab:lora_higher_rank} shows the performance of LoRA for different rank sizes. Notice that average performance, in terms of accuracy and NLE quality, do not increase when the rank is increased. 

% Table for comparison with LoRA for higher ranks
\begin{table*}[t]
    \centering
    \resizebox{\textwidth}{!}{
        \begin{tabular}{c c c c c c c c c}
            \toprule
            \shortstack{PEFT \\ Strategy} & \shortstack{Rank \\ Size} & \shortstack{Percentage \\ Parameters} & {} &  \textbf{ComVE} &                                     \textbf{ECQA} &                                     \textbf{SBIC} &                                   \textbf{e-SNLI} &                                      \textbf{Avg} \\
            \midrule
            \multirow{10}{*}{LoRA} & 8 & 0.32\% & Acc. &  67.64 {\tiny $\pm3.37$}  &  39.59 {\tiny $\pm3.82$}  &  63.42 {\tiny $\pm3.46$}  &   84.15 {\tiny $\pm2.0$}  &   63.7 {\tiny $\pm3.16$}  \\
                 &  &  & nBERTs &  61.24 {\tiny $\pm3.09$}  &   1.55 {\tiny $\pm1.26$}  &  60.93 {\tiny $\pm3.45$}  &  76.41 {\tiny $\pm1.82$}  &   50.03 {\tiny $\pm2.4$}  \\
                 & 16 & 0.63\% & Acc. &   67.94 {\tiny $\pm3.4$}  &  39.41 {\tiny $\pm3.58$}  &  63.26 {\tiny $\pm3.36$}  &  84.26 {\tiny $\pm1.88$}  &  63.72 {\tiny $\pm3.06$}  \\
                 &  &  & nBERTs &  61.51 {\tiny $\pm3.11$}  &   1.44 {\tiny $\pm1.31$}  &  60.78 {\tiny $\pm3.49$}  &   76.5 {\tiny $\pm1.71$}  &  50.06 {\tiny $\pm2.41$}  \\
                 & 32 & 1.26\% & Acc. &  67.79 {\tiny $\pm3.75$}  &  39.74 {\tiny $\pm3.85$}  &   63.5 {\tiny $\pm3.28$}  &   84.27 {\tiny $\pm1.9$}  &   63.82 {\tiny $\pm3.2$}  \\
                 &  &  & nBERTs &  61.36 {\tiny $\pm3.43$}  &   1.36 {\tiny $\pm1.18$}  &  61.01 {\tiny $\pm3.36$}  &  76.51 {\tiny $\pm1.73$}  &  50.06 {\tiny $\pm2.43$}  \\
                 & 64 & 2.49\% & Acc. &  67.65 {\tiny $\pm3.77$}  &  43.44 {\tiny $\pm3.54$}  &  63.78 {\tiny $\pm3.15$}  &  84.25 {\tiny $\pm1.91$}  &   64.86 {\tiny $\pm3.11$}  \\
                 &  &  & nBERTs &  61.31 {\tiny $\pm3.42$}  &   0.32 {\tiny $\pm0.40$}  &  61.10 {\tiny $\pm3.31$}  &  76.54 {\tiny $\pm1.73$}  &  50.01 {\tiny $\pm2.10$}  \\
                 & 128 & 4.86\% & Acc. &  67.77 {\tiny $\pm3.73$}  &  43.51 {\tiny $\pm3.57$}  &  63.57 {\tiny $\pm3.16$}  &  84.26 {\tiny $\pm1.92$}  &   64.78 {\tiny $\pm3.1$}  \\
                 &  & & nBERTs &  61.36 {\tiny $\pm3.41$}  &   0.33 {\tiny $\pm0.41$}  &  61.06 {\tiny $\pm3.29$}  &  76.49 {\tiny $\pm1.75$}  &  49.81 {\tiny $\pm2.22$}  \\
            \bottomrule
        \end{tabular}
    }
    \caption{Accuracy and NLE quality metrics for different rank sizes in LoRA. We report the average and the standard deviation over the 60 few-shot train-validation splits for the \textbf{accuracy} metric and the normalized BERTScore~(\textbf{nBERTs}). 
    % We show in \textbf{bold} the setting with the highest metric for each dataset.
    }
    \label{tab:lora_higher_rank}
\end{table*}

%%%%%%%%%%%%%%%%%%%%%%%%%%%%%%%%%%%%%%%%%%%%%%%%%%%%%%%%
%%%%%%%%%%%%%%%%%%%%%%%%%%%%%%%%%%%%%%%%%%%%%%%%%%%%%%%%
%%%%%%%%%%%%%%%%%%%%%%%%%%%%%%%%%%%%%%%%%%%%%%%%%%%%%%%%
\subsection{Explanations Shortcomings per Dataset}\label{ap:shortcomings_per_dataset}

\begin{figure}[t]
    \centering
    \includegraphics[width=0.5\textwidth]{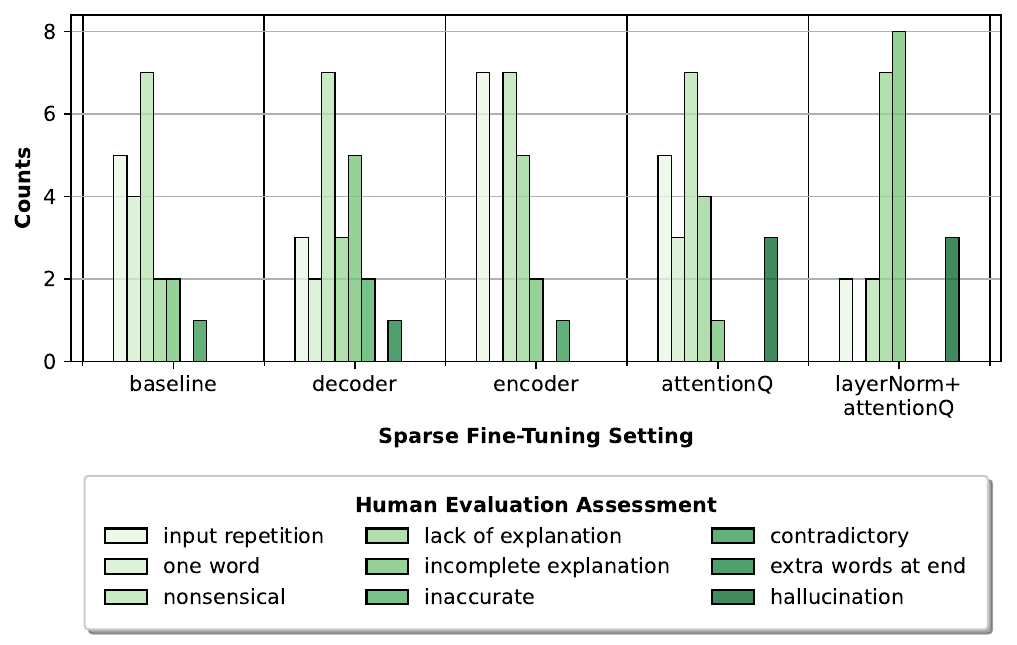}
    \caption{Histogram of the occurrences of the main shortcomings of the generated explanations for the baseline and the two best performing sparse fine-tuning setup for the \textbf{e-SNLI} dataset.}%
    \label{fig:human_plausibility_reason_esnli}
\end{figure}

Given the diverse nature of the studied datasets, we perform an individual analysis for each dataset in order to find the particular deficiencies and traits of the explanations by dataset. Figure~\ref{fig:human_plausibility_reason_esnli} shows a set of histograms with the assessment of the annotators on shortcomings for the e-SNLI dataset. It can be seen that the \textit{Nonsensical} category is consistently the most common no matter what fine-tuning strategy was used. Below, the reader can find two examples of \textit{Nonsensical} explanations generated by the \textit{Baseline} and the \textit{Decoder} strategy, respectively.

% Examples Non-sensical e-SNLI
\begin{figure}[t]
        \begin{tcolorbox}[colback=green!5!white, colframe=black,
        left=1mm,top=1mm,right=1mm,bottom=1mm, fonttitle=\small]
        \resizebox{\columnwidth}{!}{
            % Example 1
            \begin{tabular}{ p{7em} | p{22em} } 
                \textbf{Premise} &  A poor family is leaving their home with only a few belongings \\ 
                \textbf{Hypothesis} &  A man eats a chalupa \\
                \textbf{Label} & Contradiction \\
                \textbf{Explanation} & A family consists of two or more people, not just one man. \\
                &  \\
                &  \textbf{Generated Explanation}  \\
                & ``a man who eats a chalupa also has to be poor" \\
                \textbf{Human Score} & No \\
                \textbf{Reason} & Nonsensical \\
            \end{tabular}
            }
        \end{tcolorbox}
        \begin{tcolorbox}[colback=green!5!white, colframe=black,
        left=1mm,top=1mm,right=1mm,bottom=1mm, fonttitle=\small]
        \resizebox{\columnwidth}{!}{
            % Example 2
            \begin{tabular}{ p{7em} | p{22em} } 
                \textbf{Premise} &  A man in red pants skiing down a slope \\ 
                \textbf{Hypothesis} & An Olympic skier skiing. \\
                \textbf{Label} & Entailment \\
                \textbf{Explanation} & WE have no idea if the man is an olympic skier or not. \\
                &  \\
                &  \textbf{Generated Explanation}  \\
                & ``we don't know what he is doing" \\
                \textbf{Human Score} & No \\
                \textbf{Reason} & Nonsensical \\
            \end{tabular}
        }
        \end{tcolorbox}
    \caption{Examples of \textbf{Non-sensical} NLEs generated for e-SNLI.}
    \label{fig:examples_nonsensical_nle_esnli}
\end{figure}

% Plot reason e-SNLI

In addition to this, \textit{Input Repetition} is the second most common shortcoming for e-SNLI. A regular pattern found in the generated explanations is the repetition of a sub-string of the hypothesis as the predicted explanation, which happens for around $17\%$ of the generated explanations. Below, the reader can see an example of input repetition found in the e-SNLI dataset.

% Examples Input Repetition e-SNLI
\begin{figure}[t]
        \begin{tcolorbox}[colback=green!5!white, colframe=black,
        left=1mm,top=1mm,right=1mm,bottom=1mm, fonttitle=\small]
        \resizebox{\columnwidth}{!}{
            % Example 1
            \begin{tabular}{ p{7em} | p{22em} } 
                \textbf{Premise} &  girl in uniform running through the water fountain gushing water from the tiles \\ 
                \textbf{Hypothesis} & A girl is running through the water \\
                \textbf{Label} & Entailment \\
                \textbf{Explanation} & ``If a water fountain is gushing water, that implies that there is water, and therefore the girl is running through water" \\
                &  \\
                &  \textbf{Generated Explanation}  \\
                & ``a girl in uniform running through the water fountain." \\
                \textbf{Human Score} & No \\
                \textbf{Reason} & Input Repetition \\
            \end{tabular}
        }
        \end{tcolorbox}
    \caption{Examples of \textbf{Input-Repetition} NLEs generated for e-SNLI.}
    \label{fig:examples_inputrepetion_nle_esnli}
\end{figure}

\begin{figure}[t]
    \centering
    \includegraphics[width=0.5\textwidth]{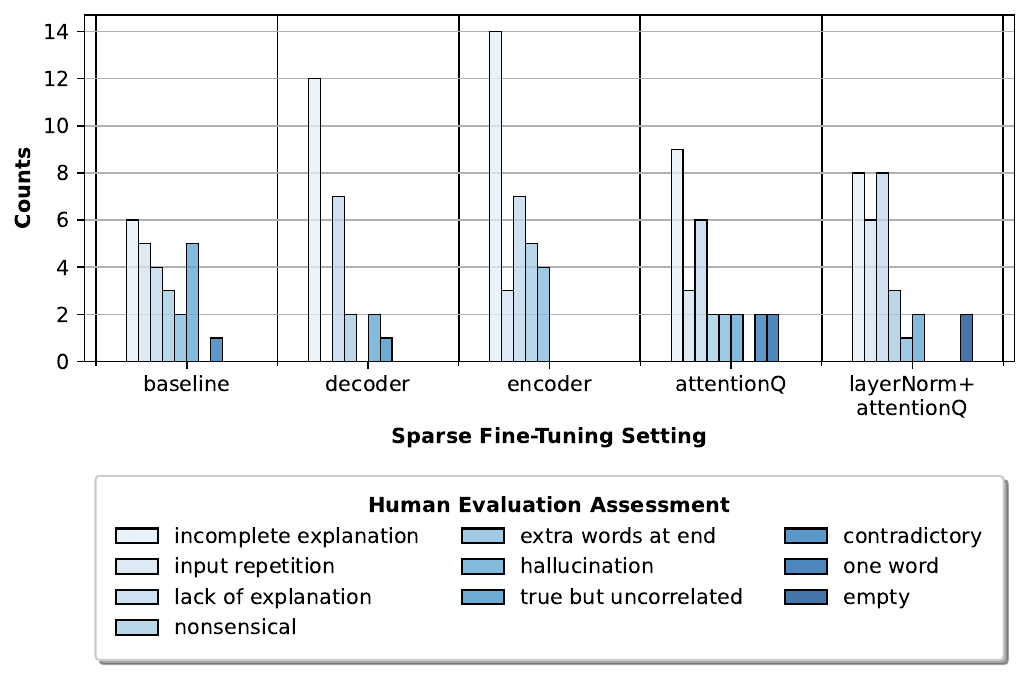}
    \caption{Histogram of the occurrences of the main shortcomings of the generated explanations for the baseline and the two best performing sparse fine-tuning setup for the \textbf{ECQA} dataset.}%
    \label{fig:human_plausibility_reason_ecqa}
\end{figure}

We depict in \cref{fig:human_plausibility_reason_ecqa} a set of histograms with the number of times that a shortcoming category happens for different fine-tuning strategies for ECQA. Predominantly, \textit{Incomplete Explanation} is the main weakness of generated NLEs. Notice that for this dataset, the answers are not generally shared by different samples (i.e., the possible labels for a sample are not always the same as in the other datasets). This causes the generated explanations to be vague and incomplete. Below, the reader can see 3 examples of \textit{Incomplete Explanation} generated by the \textit{Baseline}, \textit{Decoder}, and \textit{Encoder} fine-tuning strategy, respectively.

% Examples Incomplete Explanation
\begin{figure}[t]
        \begin{tcolorbox}[colback=blue!5!white, colframe=black,
        left=1mm,top=1mm,right=1mm,bottom=1mm, fonttitle=\small]
        \resizebox{\columnwidth}{!}{
            % Example 1
            % Example 1 
            \begin{tabular}{ p{7em} | p{22em} } 
                \textbf{Question} & Where is the ticket booth often outside the building? \\ 
                \textbf{Choices} & train station --- stadium --- auditorium --- subway  \\
                \textbf{Label} & auditorium \\
                \textbf{Explanation} & Ticket booth of an auditorium is often outside the building \\
                &  \\
                &  \textbf{Generated Explanation}  \\
                & ``auditorium is a large venue with lots of people. the ticket booth is often outside the building" \\
                \textbf{Human Score} & Weak Yes \\
                \textbf{Reason} & Incomplete Explanation \\
            \end{tabular}
        }
        \end{tcolorbox}
        \begin{tcolorbox}[colback=blue!5!white, colframe=black,
        left=1mm,top=1mm,right=1mm,bottom=1mm, fonttitle=\small]
        \resizebox{\columnwidth}{!}{
            % Example 1
            \begin{tabular}{ p{7em} | p{22em} } 
                \textbf{Question} & What happens to the person that is well known for frequent farting? \\ 
                \textbf{Choices} & social ostracism --- evil --- bad smell --- stinking \\
                \textbf{Label} & social ostracism \\
                \textbf{Explanation} & Ticket booth of an auditorium is often outside the building \\
                &  \\
                &  \textbf{Generated Explanation}  \\
                & ``Social ostracism is act of ignoring and excluding individuals. Frequent farting can make you ignored by people around you as frequent farting makes them uncomfortable with you" \\
                \textbf{Human Score} & Weak Yes \\
                \textbf{Reason} & Incomplete Explanation \\
            \end{tabular}
        }
        \end{tcolorbox}
    \caption{Examples of \textbf{Incomplete Explanation} NLEs generated for ECQA.}
    \label{fig:examples_incomplete_nle_ecqa}
\end{figure}

% Plot reason ECQA

\begin{figure}[t]
    \centering
    \includegraphics[width=0.5\textwidth]{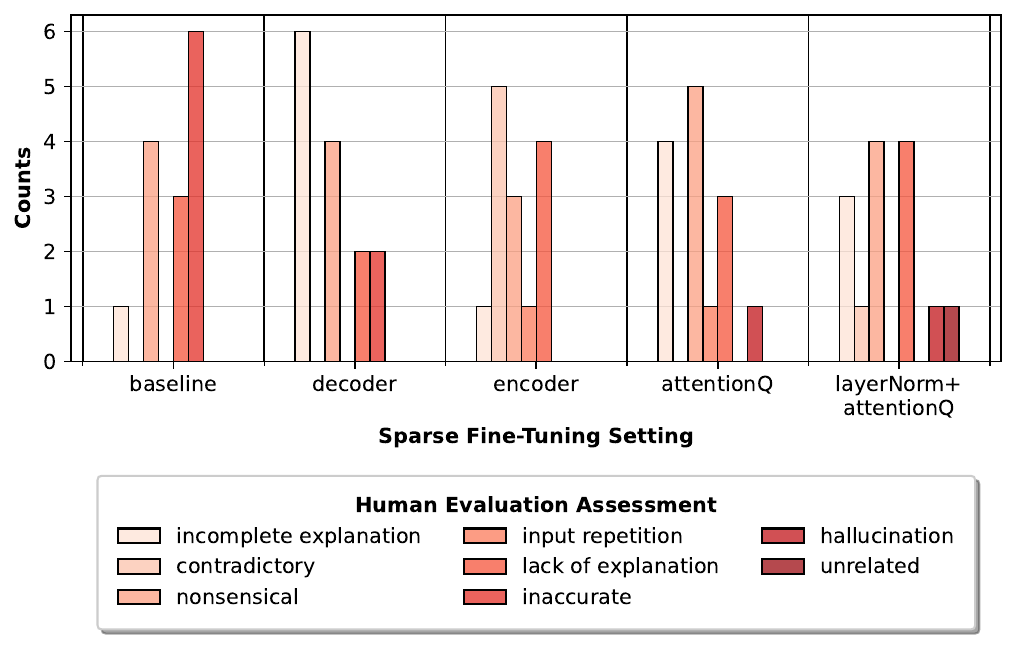}
    \caption{Histogram of the occurrences of the most common explanation shortcomings for the baseline and the two best performing sparse fine-tuning setup for the \textbf{SBIC} dataset.}%
    \label{fig:human_plausibility_reason_sbic}
\end{figure}

\cref{fig:human_plausibility_reason_sbic} shows a set of histograms with the assessment done by the annotators about the most common shortcomings. Different from other datasets, there is no singular shortcoming that dominates the results for all the fine-tuning setups. The most common mistakes among all the explanations in the dataset are: \textit{Inaccurate}, \textit{Nonsensical}, and \textit{Incomplete Explanation}. Below, the reader can find an example for the \textit{Incomplete Explanation} shortcoming for the \textit{Decoder} fine-tuning. % (\textcolor{orange}{WARNING: Due to the nature of the dataset, the content in the provided example could be offensive!}).

% Examples Incomplete Explanation SBIC
\begin{figure}[t]
        \begin{tcolorbox}[colback=red!5!white, colframe=black,
        left=1mm,top=1mm,right=1mm,bottom=1mm, fonttitle=\small]
        \resizebox{\columnwidth}{!}{
            % Example 1
            \begin{tabular}{ p{7em} | p{22em} } 
                \textbf{Post} & The normie "vibe" argument is horseshit Anytime a normie says that you give off misogynistic vibes even if you don't outright say "I hate women" they are lying. \\ 
                \textbf{Label} & Offensive \\
                \textbf{Explanation} & this post is a personal attack \\
                &  \\
                &  \textbf{Generated Explanation}  \\
                & ``this post implies that women are inferior" \\
                \textbf{Human Score} & Weak No \\
                \textbf{Reason} & Incomplete Explanation \\
            \end{tabular}
        }
        \end{tcolorbox}
    \caption{Examples of \textbf{Incomplete Explanation} NLEs generated for SBIC.}
    \label{fig:examples_incomplete_nle_sbic}
\end{figure}

% Plot reason SBIC

\begin{figure}[t]
    \centering
    \includegraphics[width=0.5\textwidth]{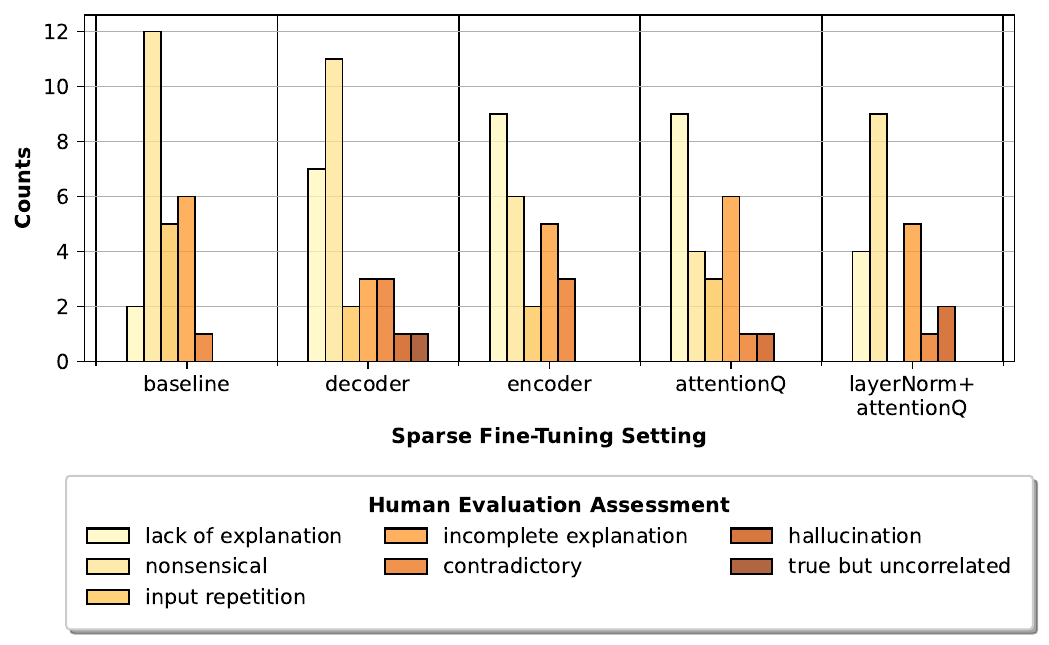}
    \caption{Histogram of the occurrences of the most common explanation shortcomings for the baseline and the two best performing sparse fine-tuning setup for the \textbf{ComVE} dataset.}%
    \label{fig:human_plausibility_reason_comve}
\end{figure}

We have depicted in \cref{fig:human_plausibility_reason_comve} a series of histograms with the frequency of possible shortcomings given by human annotators to the evaluated explanations. It can be seen that annotators consider that the \textit{Lack of explanation}, \textit{Nonsensical}, and \textit{Incomplete Explanation} are the most relevant categories to explain the weaknesses of the generated explanations. 

% Plot reason ComVE

\subsection{Inter-annotator Agreement}\label{ap:examples_inter_agreement}

We show in \cref{fig:examples_perceptul_disagreement} an example of perceptual disagreement where the annotators assigned the same plausibility reason but a different score. Furthermore,~\cref{fig:examples_expectation_disagreement_end} shows an example of expectation disagreement where human evaluators assigned a opposite score for the given explanation.

% Expectation disagrement example
\begin{figure}[t]
        \begin{tcolorbox}[left=1mm,top=1mm,right=1mm,bottom=1mm, colback=blue!5!white, colframe=black]
        \centering
        \resizebox{\columnwidth}{!}{
            \begin{tabular}{ p{7em} | p{22em} } 
                \textbf{Question} & What is the best way to release energy? \\ 
                \textbf{Choices} & yell --- think --- exercise --- rest --- work off   \\
                \textbf{Label} & exercise \\
                \textbf{Explanation} & By doing exercise, one can release energy. The best way of releasing energy is exercise. \\
                % &  \\
                &  \textbf{Generated Explanation}  \\
                & ``exercise releases energy. to release energy, one must yell or yell loudly. to exercise, one must exercise vigorously. to work off, one must work off the energy." \\
                % & \\
            \end{tabular}
        }
        \\
        \resizebox{\columnwidth}{!}{
            \begin{tabular}{p{6.66em}|p{10em}|p{10em}}
                 & \textbf{Annotator 1} & \textbf{Annotator 2} \\ \hline
                 \textbf{Human Score } & No & Weak Yes \\
                 \textbf{Reason} & Hallucination & Hallucination \\
            \end{tabular}
            }
    \end{tcolorbox}
    \caption{Example of annotator expectation disagreement in our study for the ECQA dataset.}
    \label{fig:examples_expectation_disagreement}
\end{figure}

% Perceptual disagreement
\begin{figure}[t]
        \begin{tcolorbox}[left=1mm,top=1mm,right=1mm,bottom=1mm, colback=blue!5!white, colframe=black]
        \centering
        \resizebox{\columnwidth}{!}{
        \begin{tabular}{ p{7em} | p{22em} } 
            \textbf{Question} & Teddy liked learning languages.  He helped him with what? \\ 
            \textbf{Choices} & problems --- frustration --- confidence --- better communication --- sadness   \\
            \textbf{Label} & better communication \\
            \textbf{Explanation} & Better communication is defined as verbal speech or other methods of relaying information that get a point across. % Teddy liked learning languages. 
            He helped him with better communication. \\
            &  \\
            &  \textbf{Generated Explanation}  \\
            & he helped him with better communication. \\
            & \\
        \end{tabular}
        }
        \\
        \resizebox{\columnwidth}{!}{
        \begin{tabular}{p{6.66em}|p{10em}|p{10em}}
             & \textbf{Annotator 1} & \textbf{Annotator 2} \\ \hline
             \textbf{Human Score } & Weak No &  Weak Yes \\
             \textbf{Reason} & Lack of Explanation & Lack of Explanation \\
        \end{tabular}
        }
        \end{tcolorbox}
    %\vspace{-1em}
    \caption{Example of annotator perceptual disagreement in our study for the ECQA dataset.}
    \label{fig:examples_perceptul_disagreement}
\end{figure}

% Expectation disagrement example
\begin{figure}[t]
        \begin{tcolorbox}[left=1mm,top=1mm,right=1mm,bottom=1mm, colback=blue!5!white, colframe=black]
        \centering
        \resizebox{\columnwidth}{!}{
            \begin{tabular}{ p{7em} | p{22em} } 
                \textbf{Question} & What is the best way to release energy? \\ 
                \textbf{Choices} & yell --- think --- exercise --- rest --- work off   \\
                \textbf{Label} & exercise \\
                \textbf{Explanation} & By doing exercise, one can release energy. The best way of releasing energy is exercise. \\
                &  \\
                &  \textbf{Generated Explanation}  \\
                & ``exercise releases energy. to release energy, one must yell or yell loudly. to exercise, one must exercise vigorously. to work off, one must work off the energy." \\
                & \\
            \end{tabular}
        }
        \\
        \resizebox{\columnwidth}{!}{
            \begin{tabular}{p{6.66em}|p{10em}|p{10em}}
                 & \textbf{Annotator 1} & \textbf{Annotator 2} \\ \hline
                 \textbf{Human Score } & No & Weak Yes \\
                 \textbf{Reason} & Hallucination & Hallucination \\
            \end{tabular}
            }
    \end{tcolorbox}
    \caption{Example of annotator expectation disagreement in our study for the ECQA dataset.}
    \label{fig:examples_expectation_disagreement_end}
\end{figure}

\end{document}